\documentclass{article}
\usepackage{ijcai20}

\usepackage{subfiles} 
\usepackage[none]{hyphenat}

\usepackage{times}

\usepackage{soul}
\usepackage{url}
\usepackage{enumitem}
\usepackage[normalem]{ulem}

\PassOptionsToPackage{hyphens}{url}
\usepackage[colorlinks = true,
            linkcolor = red,
            urlcolor  = red,
            citecolor = red,
            anchorcolor = red]{hyperref}
\expandafter\def\expandafter\UrlBreaks\expandafter{\UrlBreaks
\do\a\do\b\do\c\do\d\do\e\do\f\do\g\do\h\do\i\do\j%
\do\k\do\l\do\m\do\n\do\o\do\p\do\q\do\r\do\s\do\t%
\do\u\do\v\do\w\do\x\do\y\do\z\do\A\do\B\do\C\do\D%
\do\E\do\F\do\G\do\H\do\I\do\J\do\K\do\L\do\M\do\N%
\do\O\do\P\do\Q\do\R\do\S\do\T\do\U\do\V\do\W\do\X%
\do\Y\do\Z}

\usepackage[utf8]{inputenc}
\usepackage[small]{caption}
\usepackage{graphicx}
\usepackage{subcaption}
\usepackage{caption}
\usepackage{amsmath}
\usepackage{booktabs}
\usepackage{float}      
\usepackage{abstract}
\usepackage{xcolor}
\usepackage{adjustbox}
\usepackage{tabu} 
\usepackage{multicol}
\usepackage{cuted} 
\usepackage{caption} 
\usepackage{pdfpages}

\usepackage{array}
\usepackage{rotating}

\usepackage[sorting=none]{biblatex}
\title{Identifying Ethical Biases in Action Recognition Models}

\author{
Ana Băltărețu$^1$
\and
Pascal Benschop$^1$
\and
Jan van Gemert$^1$
\affiliations
$^1$EEMCS, Delft University of Technology, The Netherlands
\emails
\{a.baltaretu, p.benschop, j.c.vangemert\}@tudelft.nl
}

\begin{document}


\maketitle/

\begin{strip}
  \centering
    \vspace*{-1cm}

  \rotatebox{90}{\makebox[8em][c]{\textbf{Video 1}}}
  \includegraphics[width=0.95\textwidth]{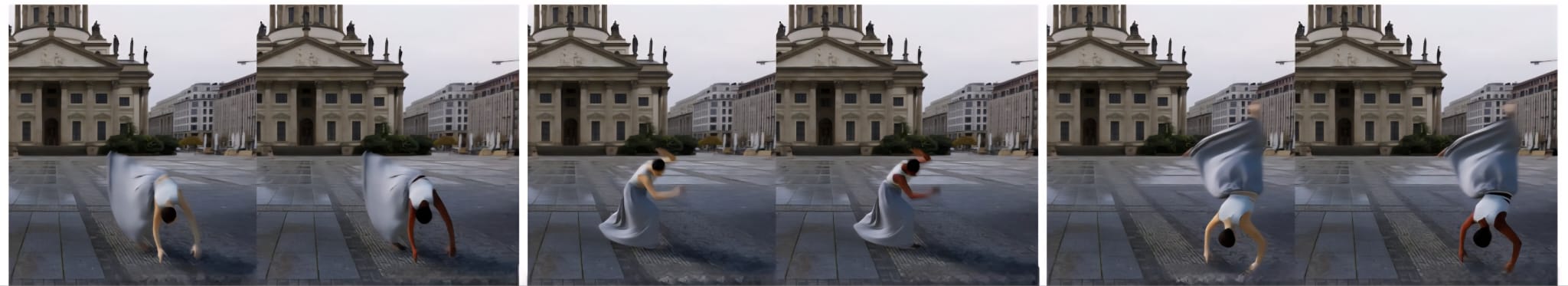}  
  \vspace*{-0.1cm}
  
{\scriptsize
Left: lighter skin tone, right: darker skin tone, prediction change from ``cartwheeling'' to ``capoeira''.
} \vspace{0.1cm}

  \rotatebox{90}{\makebox[8em][c]{\textbf{Video 2}}}
  \includegraphics[width=0.95\textwidth]{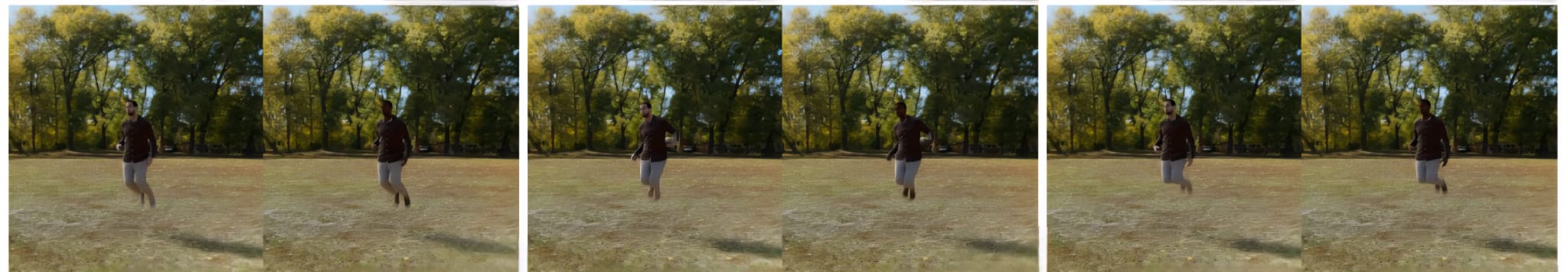}
  \vspace*{-0.1cm}
  
{\scriptsize
Left: lighter skin tone, right: darker skin tone, prediction change from ``jumpstyle dancing'' to ``juggling soccer ball''. 
}

\vspace*{-0.1cm}

  \captionof{figure}{Qualitative analysis  showcasing potential racial bias in action recognition models. Predicted labels per video at the bottom right of the frame. In Video 1 (top row), all sequences depict the same motion performed in the same environment, with only the actor’s skin tone changed. The model correctly labels the lighter-skinned person as ``cartwheeling'', while misclassifying the darker-skinned person as performing ``capoeira'' an Afro-Brazilian martial art that blends dance and acrobatics. In Video 2 (bottom row), similar inconsistencies occur for the action ``jumpstyle dancing'' depending on the perceived race of the actor. See: \url{https://youtu.be/amygAq-Sqc4}. This suggests that some action predictions rely on visual attributes rather than on the movement.}
  \label{fig:qualitative_analysis}
\end{strip}

\begin{abstract}
Human Action Recognition (HAR) models are increasingly deployed in high-stakes environments, yet their fairness across different human appearances has not been analyzed. 
We introduce a framework for auditing bias in HAR models using synthetic video data, generated with full control over visual identity attributes such as skin color. 
Unlike prior work that focuses on static images or pose estimation, our approach preserves temporal consistency, allowing us to isolate and test how changes to a single attribute affect model predictions. 
Through controlled interventions using the BEDLAM simulation platform, we show whether some popular HAR models exhibit statistically significant biases on the skin color even when the motion remains identical. 
Our results highlight how models may encode unwanted visual associations, and we provide evidence of systematic errors across groups. 
This work contributes a framework for auditing HAR models and supports the development of more transparent, accountable systems in light of upcoming regulatory standards.






\vspace{0.2cm}
\end{abstract}
\section{Introduction}

Can an action recognition model mistake a cartwheel for capoeira simply because the actor has a darker skin tone? Such questions are central to evaluating fairness in Human Action Recognition (HAR), where predictions may reflect underlying bias, see \autoref{fig:qualitative_analysis}.
We present a framework for auditing fairness in Human Action Recognition (HAR) models using synthetic data.
By systematically varying visual attributes our approach exposes potential biases in model predictions that may otherwise remain hidden in real-world datasets. 
This is increasingly relevant as HAR models are integrated into high-stake domains such as security, autonomous driving, healthcare \cite{zhao2024review}, and regulatory frameworks like the EU AI Act \cite{down2021proposal, edwards2021eu} demand greater transparency and accountability from developers of AI systems.
With this paper we are trying to assist policy makers, law enforcers and AI engineers in developing more robust and fair systems, making it easier to verify their compliance with emerging legal and ethical standards.
To this end, we developed a framework for generating synthetic video datasets with complete control over a subset of visual attributes. 
Modifying a single attribute allows us to isolate their effects and measure bias without introducing confounding factors.

\textit{HAR models exhibit bias because they do not always learn to recognize actions based purely on human movement.} Instead, they may rely on visual attributes that are not directly related to the action itself, such as clothing, background, or body appearance \cite{ilic2022appearance, vu2014predicting, he2016human, choi2019can, chung2022enabling}. For instance, a model might associate a Hawaiian shirt with drinking a cocktail, even though the shirt has nothing to do with the action being performed. This suggests that the model could be picking up on superficial visual patterns in the training data, rather than truly understanding the motion, which raises concerns about fairness and reliability.

In this project, we identify how sensitive models are to changing ethical attributes. 
Regulatory frameworks like the EU AI Act \cite{down2021proposal, edwards2021eu} highlight risks such as opacity and data dependency, which can lead to discriminatory outcomes in AI systems. 
Motivated by these concerns, we focus on potential biases linked to race or skin color, 
which may undermine fundamental rights like non-discrimination \cite{europeanunion_2000}. 
Our analysis does not include general action recognition tasks like industrial processes, or animal monitoring, instead, due to the inherently human nature of our selected attributes, we limit the analysis to the HAR domain.

Some approaches for evaluating biases have been to manually annotate \cite{buolamwini2018gender}, or derive information from existing labels \cite{wang2019balanced}, and then assess model performance for each group. 
An emerging trend has been to generate controlled synthetic images using neural networks, for facial recognition \cite{liang2023benchmarking}, or human pose and shape estimation \cite{stage}.
Similarly, for the task of Human Action Recognition we require video data in which specific attributes can be controlled.
While prior work such as STAGE \cite{stage} has demonstrated bias analysis in individual frames, HAR introduces an additional challenge: motion across frames.
This requirement makes extending frame-level approaches like STAGE to the video domain particularly challenging and prone to generation defects, resulting in less control. 

BEDLAM \cite{bedlam} allows us to generate physically plausible and temporally consistent videos while programmatically controlling visual attributes. 
To avoid introducing confounding factors because of potential artifacts, we adopt the fully simulated approach through BEDLAM.
The simulated approach enables isolated interventions such as changing skin tone, so we can systematically evaluate how these attributes influence model predictions.


This paper aims to investigate biases from human action recognition (HAR) models in a simulated environment. Specifically, it addresses the main research question:
\textit{\textbf{How sensitive are Human Action Recognition (HAR) models to single visual attribute changes in synthetically generated RGB videos?}}
To explore this question, the paper focuses on two sub-research questions:
\begin{enumerate}[leftmargin=*, label=SubQ\arabic*.]
    \item 
    How well do HAR models trained on real data generalize to synthetic videos?

    \item 
    How can we measure significant differences in HAR predictions when changing personal characteristics?

\end{enumerate}

\noindent To address the research questions, we present the following contributions: we evaluated publicly available action recognition models directly on fully synthetic data without fine-tuning, observing that while some models under-perform, others yield promising results. 
Finally, we focused on skin color as a key parameter to investigate model bias, identifying cases where predictions shift significantly with changes in appearance.

\section{Related work}
\label{Related work}





Past research has been exploring biases across various domains of deep learning. 
This includes studies on gender bias in text-to-image generative models \cite{girrbach2025large}, 
and visual bias in vision-language models \cite{huang2025visbias}. 
Within the field of action recognition, prior research has investigated model behavior under occlusion \cite{grover2023revealing}, 
as well as their robustness to a wide range of noise types \cite{schiappa2023large}. 
Our paper focuses on making a framework for auditing action recognition models and the datasets they are trained on. 
We want to identify if they have any inherent biases towards ethical attributes such as skin color. 
Previously there have been other papers who made similar auditing, testing and diagnostics frameworks
\cite{stage, duran2025metamorphic, wang2025pulsecheck457}.
STAGE \cite{stage} focuses on the task of Human Pose and Shape estimation (HPS) models, while \cite{schiappa2023large} focuses on robustness analysis on 90 perturbations (e.g. noise, blur, camera movement). 
At their core these tasks are similar to action recognition, but operate on individual frames, while action recognition requires reasoning over entire motion sequence.



\subsection{What makes HAR so challenging?}
\label{subsec:HAR_so_challenging}
Human Action Recognition (HAR) using RGB monocular data is a complex task due to the large variety of conditions under which actions occur.
Actions are frequently associated with specific environments, camera angles, lighting conditions, and backgrounds \cite{moon2021integralaction}. 
This association is likely due to their \textbf{inter-class variability}, since videos of actions under the same label occur in similar settings. 
For instance, actions like ``playing football'' typically take place in outdoor fields under natural lighting, whereas ``cooking'' is performed in a crowded indoors space with artificial lighting. 
Actions appear different from varying viewpoints \cite{liu2017viewpoint}, so many models assume a fixed viewpoint \cite{ramanathan2014human, rawat2021view}.
This strong correlation between actions and their typical visual contexts can lead models to overfit on these appearance cues, reducing their generalizability across diverse scenarios.

We identified \textbf{viewpoint} and \textbf{background} as important attributes that do not have an obvious default value, and that have a major impact on performance when generating synthetic data.
Previous works have explored the effects of changing the camera angle \cite{varol2021synthetic}, 
as well as adding a loss metric for the background \cite{choi2019can}, but they both use synthetic data for training the models instead of testing. 
In our own experiments, we also observed that the viewpoint and background changes significantly impact model performance, which is why we chose to control these factors manually, and we limit their impact through ablation studies.

Human Action Recognition (HAR) faces significant challenges due to \textbf{intra-class variability} \cite{camarena2023overview, zunino2017revisiting, ferrari2020personalization}. 
This variability arises because individuals perform the same action differently: variations in speed, style, and body movements are common. 
Even the same person may not fully replicate an action across multiple instances \cite{ferrari2020personalization}. 
The lack of replicability introduces unwanted confounding factors, which affect measurements of independent variables.
This makes it difficult to conduct meaningful analyses of attribute changes in real (recorded, non-synthetic) datasets.  
Therefore, our study leverages synthetic data to systematically examine these specific variations under controlled conditions, by replicating the exact motion over multiple videos.

\subsection{GenAI vs Recorded vs Simulated benchmarks}


With the rise of video generation models like Veo 3 \cite{veo_2025}, as well as no-longer-state-of-the-art approaches like SORA \cite{brooks2024video} and other recent methods \cite{runway_gen4, polyak2025moviegencastmedia}, you can't help but wonder if such tools could be used to generate video data. 
Although producing realistic videos has traditionally been challenging, due to issues with temporal coherence and the introduction of visual artifacts, Veo 3 demonstrates how rapidly these models are evolving, showing noticeably fewer glitches than earlier approaches. 
This progress suggests that we are approaching a point where generating visually coherent videos may become feasible. 
However, our goal requires generating multiple videos that differ by an individual attribute, and this level of controlled consistency is unlikely to be achievable with diffusion-based models, given their potential for introducing artifacts. 
Therefore, we will not be using generative AI for creating our data.

Another way of auditing would be through recorded videos of people with different visual attributes performing the same action. 
A downside of the recorded approach is that human actors introduce uncontrollable variation \cite{zhang2019comprehensive} (e.g. timing, joint positioning, or scene lighting), which constitute confounding factors. 
By contrast, synthetic data ensures perfect reproducibility, allowing us to pinpoint model behavior under precisely defined conditions. 
This also makes the influence of the independent variable that we alter (skin tone) measurable.
Furthermore, synthetic environments are highly scalable and can be adapted retroactively (e.g., changing camera angles or background) without requiring entirely new data collection efforts, thus we prefer the simulated approach.

We generate our synthetic dataset through a simulated approach, using BEDLAM \cite{bedlam}.
A recent trend in Computer Vision has been using simulations to generate a large amount of high quality training data \cite{griffiths2019synthcity, xiang2021taking, wood2021fake, hasson2019learning, jang2020etri}. 
This approach has proven to be highly effective for training models, that are then tested on real data. 
The good performance has made us question whether the inverse method is possible.
Hence, after training the models on real data, we are testing them on synthetic data.




\subsection{The golden goose: BEDLAM}

BEDLAM \cite{bedlam} showed that training human-pose-and-shape estimation models entirely on synthetic images can still achieve state-of-the-art performance on real data. 
Their method makes sense for their work because adopting the simulated route made the annotation process simpler, whereas annotating full bodies is a tedious and error-prone process.
Although label construction is much easier in our action-recognition study, we use BEDLAM's robust simulation pipeline for a different objective: we begin with models that have been pretrained on real videos and evaluate their resilience on a synthetic benchmark.
In other words, where BEDLAM relied on synthetic data for training, we reverse the setting to a controlled, scalable evaluation environment for real-world models.
We use the BEDLAM rendering framework\footnotemark\footnotetext{BEDLAM render repository:\url{https://github.com/PerceivingSystems/bedlam_render}} to generate our synthetic videos. 
BEDLAM makes the generation of the dataset relatively simple, since we do not have to worry about the variability of body models, clothing and animations. 
Without BEDLAM, achieving realism would have been difficult.

SMPL \cite{SMPL2015} is a model for representing realisting 3D human body meshes.
Given joint positions along with an input mesh, SMPL can predict vertex positions of the skinned characters with the help of learned blend shapes. SMPL has been widely used in recent papers \cite{stage, varol2021synthetic, babel, schneider2024synthact, meshcapade} related to synthetic human data because it is more accurate than previous human body models, and compatible with rendering software like Unreal Engine.
SMPL-X \cite{SMPL_X2019} is a more expressive version of SMPL, in terms of hands and face movements. This expressivity improvement is relevant for us when generating actions such as ``drinking'' since they rely on more precise movements of the body which would not have been possible with just SMPL. BEDLAM uses SMPL-X to generate realistic 3D body movement.



BEDLAM uses animations from AMASS \cite{AMASS} to generate realistic videos of synthetic humans. 
AMASS is a large-scale collection that merges multiple motion-capture datasets under the unified SMPL model.
Conveniently for us, each AMASS sequence comes with an associated action label.
Even though AMASS is a large dataset, the scope of action recognition models is broad, and, while good for human pose estimation tasks, the limited amount of motions could constrain future work for action recognition.
We avoid this limitation by generating videos only for action labels that match semantically in both AMASS and the dataset the models are trained on.



\subsection{Evaluated HAR models}
\label{subsec:evaluated_har_models}
In this paper we test models, but to an extent we also test the datasets that they were trained on. Models trained on monocular RGB video data tend to learn visual cues \cite{ilic2022appearance, vu2014predicting, choi2019can}. Therefore, to get comparable results, we test models pretrained on the same dataset.
Kinetics-400 \cite{kinetics} is a dataset that has a large amount of pretrained action recognition model weights \cite{feichtenhofer2019slowfast, fan2021multiscale, kim2024leveraging, feichtenhofer2020x3dxs} available\footnotemark\footnotetext{Gluon models pretrained on Kinetics-400: \url{https://cv.gluon.ai/model\_zoo/action\_recognition.html?\#kinetics400-dataset}}.
Because of this and the scale of Kinetics-400, for our experiments, we evaluated models pretrained on it.

We evaluated open-source foundation models, which makes them likely to be used in real-life applications. We did not re-train these models, instead we took models pre-trained on the Kinetics-400 dataset. This choice keeps our method future-proof and gives ``deployers'' \cite{edwards2021eu} a practical way to check their models before release. Each model had to perform single-label human-action recognition, and we needed at least a rough understanding of the training data to create representative synthetic videos. Using this criterion, we analyzed five models, out of which only three achieved sufficient accuracy on the synthetic data to make subsequent bias analysis worthwhile, as low-performing models would reduce the insight gained from the bias tests because of near-random guessing. 

\textbf{MViT} \cite{fan2021multiscale} combines the idea of features pyramids with vision transformers, enabling the model to recognize actions at varying scales. MViT is well-suited for analysis because it can recognize both large-scale movements like cartwheeling and small-scale actions like crying. 

\textbf{SlowFast} \cite{feichtenhofer2019slowfast} couples a low-frame-rate pathway that learns detailed spatial semantics with a high-frame-rate pathway that focuses on subtle, rapid motion. Fusing the two paths enables understanding across temporal scales, from sustained actions like yoga, to rapid movements like a lunge. 

\textbf{TC-CLIP} \cite{kim2024leveraging} compresses the most informative patches in each clip into a few Temporal Context (TC) tokens. It then feeds these tokens to the Video-Conditional Prompt (VP) module, giving both vision and language branches a shared clip-level memory that excels on long or partly off-screen actions such as the jog videos in our synthetic dataset.

We also tested \textbf{X3D} \cite{feichtenhofer2020x3dxs} and \textbf{Slow} \cite{feichtenhofer2019slowfast}, but their near-random performance on the baseline synthetic dataset led us to drop them from the bias analysis.


\section{Methodology: Controlled Bias Auditing}
\label{sec:Methodology}

This study presents a proof-of-concept methodology for auditing bias in Human Action Recognition (HAR) systems using synthetic video data. 
We aim to evaluate whether publicly released HAR models behave robustly when exposed to controlled variations in monocular RGB video inputs, as might occur in real-world settings \cite{kinetics, UCF101, gu2018ava, hmdb51, goyal2017something, caba2015activitynet, damen2022rescaling}. 
With the setup explained in this section, we aim to test whether synthetic data can serve as a reliable tool for isolating and evaluating potential biases in model predictions.

Framing the problem as a controlled intervention, our approach enables systematic bias evaluation in HAR systems with minimal confounding factors. 
In order to validate the use of synthetic data as a viable testing ground, we first evaluate whether action recognition models can produce correct predictions on fully synthetic video data.
To ensure that any observed differences in model predictions stem solely from the manipulated attribute, we fix all other variables: the action performed, the environment, the camera position, the clothing, and all other visual attributes related to the actor's features, while one specific attribute (skin tone) is varied. 
If a model changes its prediction when only the actor’s skin tone changes, we flag this as potential evidence of biased behavior.
This setup enables us to isolate and evaluate the influence of individual attributes on model predictions, laying the foundation for a replicable bias auditing framework. 



\subsection{Key design principles}
\label{subsec:key_design_principles}


Human Action Recognition (HAR) is a field that encompasses a large variety of actions categories and we cannot realistically check all possible action types. 
Our focus was to generate data similar to the Kinetics-400 \cite{kinetics} dataset, but it is entirely possible that a similar approach can be applied to other datasets. 
Therefore we describe a framework that can be used to generate similar data for other use cases.


To evaluate models without fine-tuning, we generate synthetic data that is similar to Kinetics-400, while remaining fully controllable and replicable. 
Real-world datasets contain too much uncontrolled variation, and current generative methods introduce artifacts that interfere with controlled experiments. 
Instead, we simulate realistic videos where only specific variables are changed. 
To keep the analysis interpretable, we constrain the dataset along three dimensions: types of actions, actor attributes, and scene appearance. This design enables targeted evaluations without overwhelming complexity.

\textbf{Action constraints.}
We generate a controlled dataset where each video features a single actor performing a single action. 
This setup is meant to be the simplest way forward, since it is compatible with a broader range of models, including those designed for multi-person scenarios, while avoiding the need to adjust models that expect simpler inputs.
We generate 4–10 second clips (common in existing HAR datasets \cite{kinetics, UCF101, hmdb51}) to reduce computational demands and isolate key variables. 
This project is a proof-of-concept, so we prioritize simplicity over modeling complex multi-actor scenes.
The types of actions are limited to all the labels provided by BEDLAM, and in our case for Kinetics \cite{kinetics} that works out well because they have similar labels, and, for other benchmark datasets like UCF101 \cite{UCF101} and HMDB51 \cite{hmdb51} there are also matching labels with BEDLAM. 
By constraining our dataset to short, single-actor actions captured from a fixed camera, we strike a balance between experimental control and broad applicability, allowing us to focus on the core objective of evaluating ethical biases in action recognition systems without introducing unnecessary complexity, while generating realistic data.

\textbf{Visual characteristics constraints.}
Each synthetic video includes seven versions of the same action, differing only in the actor’s skin texture, based on Meshcapade’s seven-category classification \cite{meshcapade}. This allows us to isolate the effect of skin tone on model predictions through controlled interventions.
We use a fixed camera to ensure that the only motion comes from the actor. Initial tests showed that background and camera angle affect recognition, likely due to visual similarities within action class. We ran ablations to select the best-performing viewpoint and background for each action.
This minimalist setup not only simplifies interpretation but also lays the groundwork for future studies involving more variation, such as clothing, body type, or dynamic cameras.

\begin{figure}

\setlength{\tabcolsep}{0.03em}
\centering

\tiny
\begin{tabular}{c@{\hspace{0.2em}}|@{\hspace{0.2em}}ccc@{\hspace{0.2em}}|@{\hspace{0.2em}}ccc}
\toprule
\raisebox{-1em}{\textbf{Skin Tone}}
& \multicolumn{3}{c}{\textbf{Near}} 
& \multicolumn{3}{c}{\textbf{Far}} \\
& \textbf{Autumn} & \textbf{Konzerthaus} & \textbf{Stadium}
& \textbf{Autumn} & \textbf{Konzerthaus} & \textbf{Stadium} \\
\midrule

\raisebox{2em}{\textbf{White}}
& \includegraphics[width=0.14\linewidth]{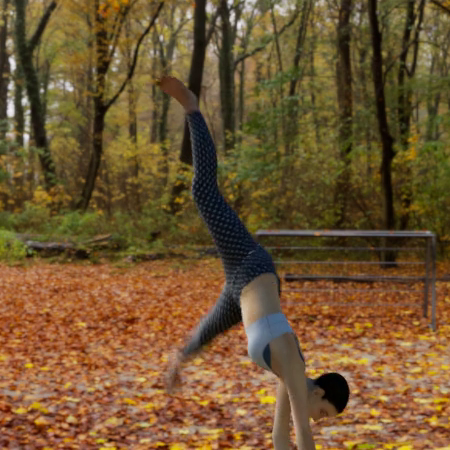}
& \includegraphics[width=0.14\linewidth]{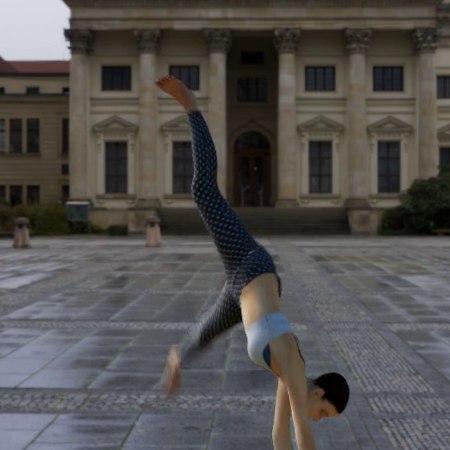}
& \includegraphics[width=0.14\linewidth]{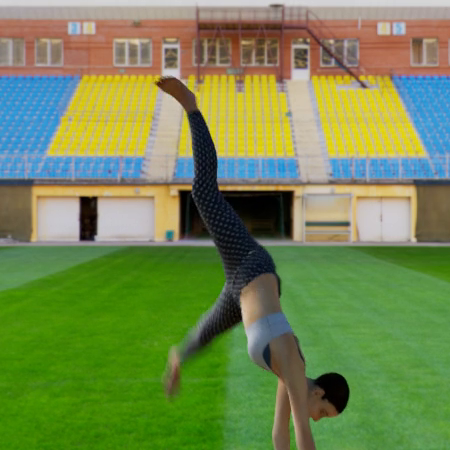}
& \includegraphics[width=0.14\linewidth]{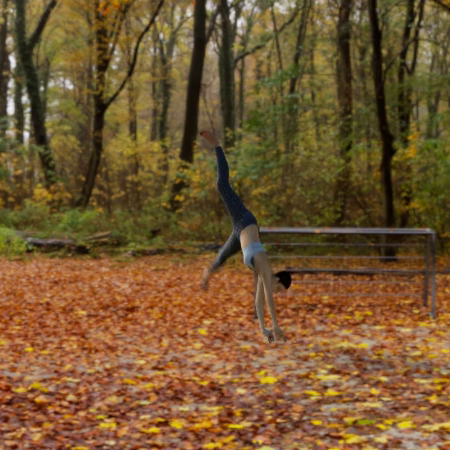}
& \includegraphics[width=0.14\linewidth]{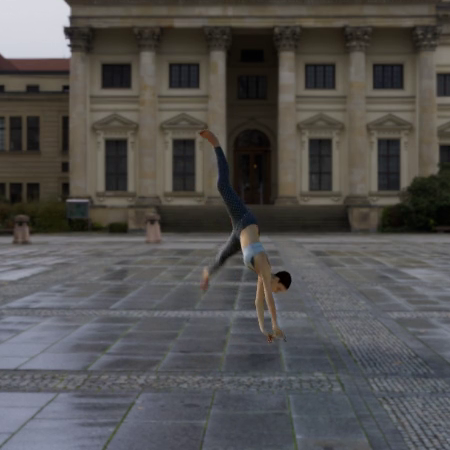}
& \includegraphics[width=0.14\linewidth]{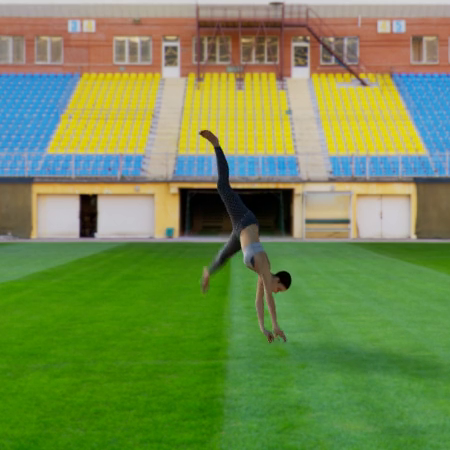} 
\vspace*{-0.2em}
\\

\raisebox{2em}{\textbf{African}}
& \includegraphics[width=0.14\linewidth]{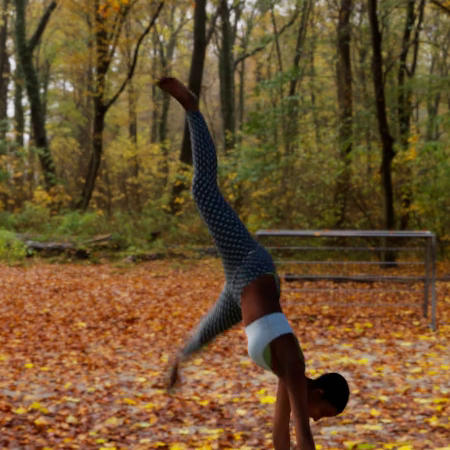}
& \includegraphics[width=0.14\linewidth]{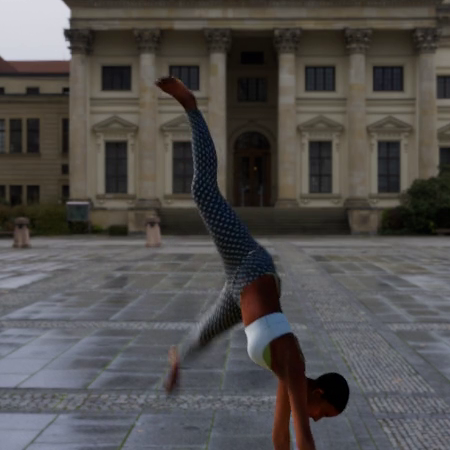}
& \includegraphics[width=0.14\linewidth]{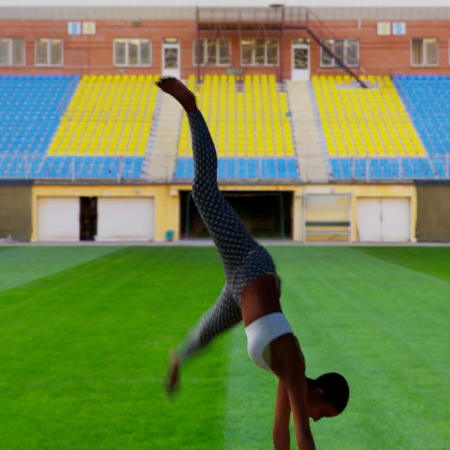}
& \includegraphics[width=0.14\linewidth]{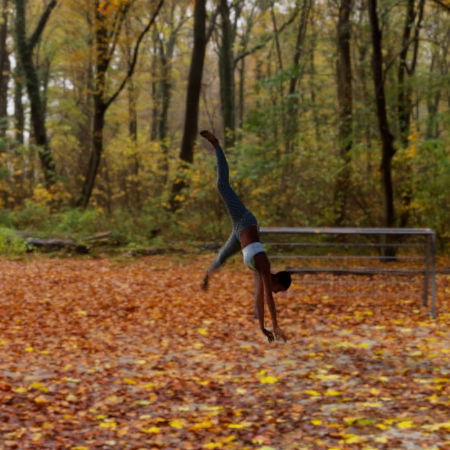}
& \includegraphics[width=0.14\linewidth]{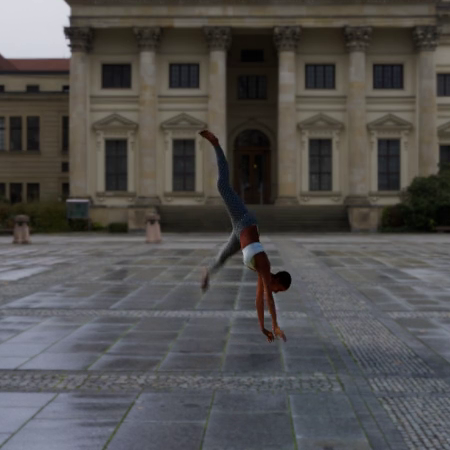}
& \includegraphics[width=0.14\linewidth]{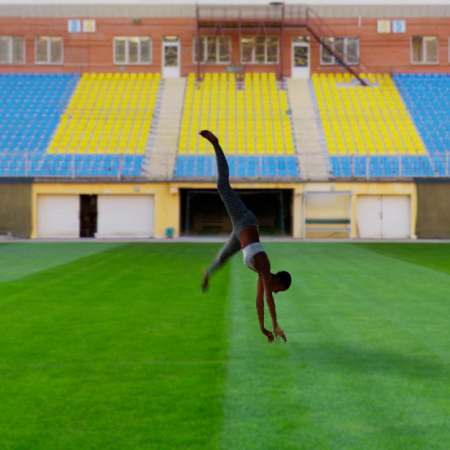} 
\vspace*{-0.2em}
\\

\raisebox{2em}{\textbf{Asian}}
& \includegraphics[width=0.14\linewidth]{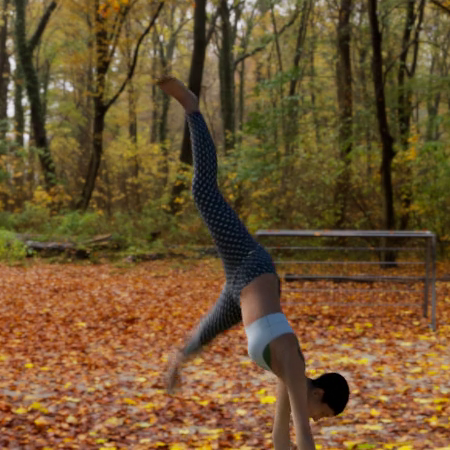}
& \includegraphics[width=0.14\linewidth]{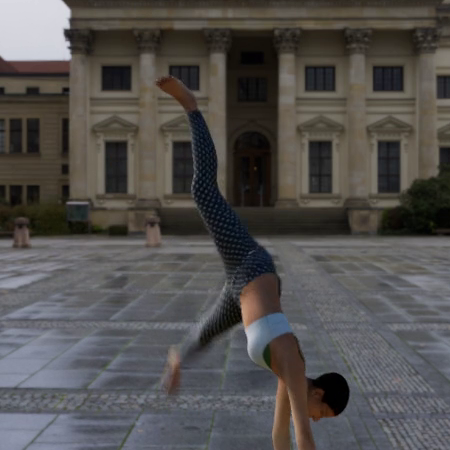}
& \includegraphics[width=0.14\linewidth]{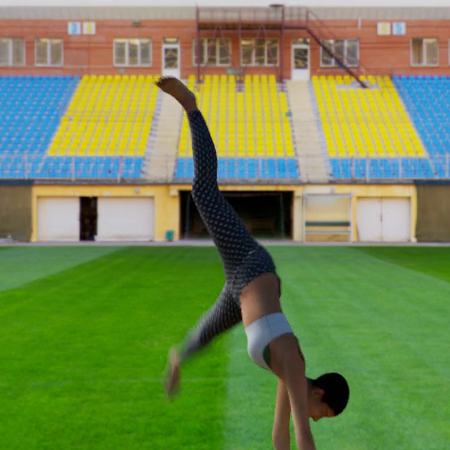}
& \includegraphics[width=0.14\linewidth]{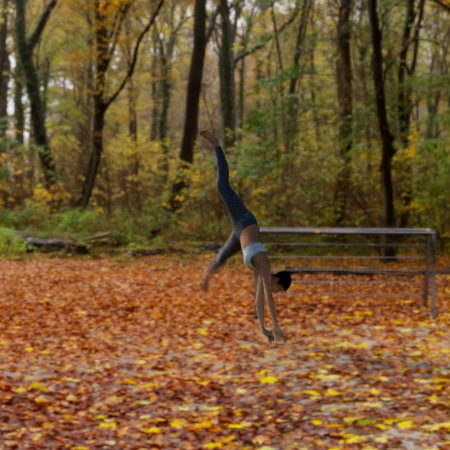}
& \includegraphics[width=0.14\linewidth]{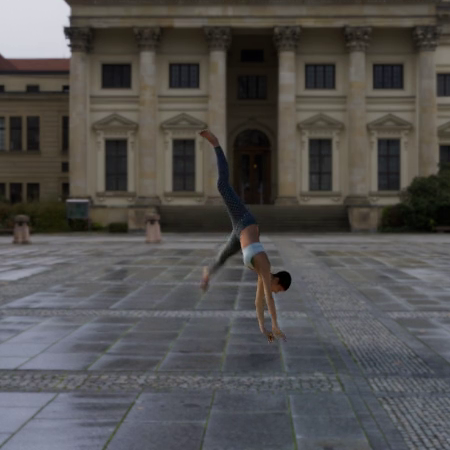}
& \includegraphics[width=0.14\linewidth]{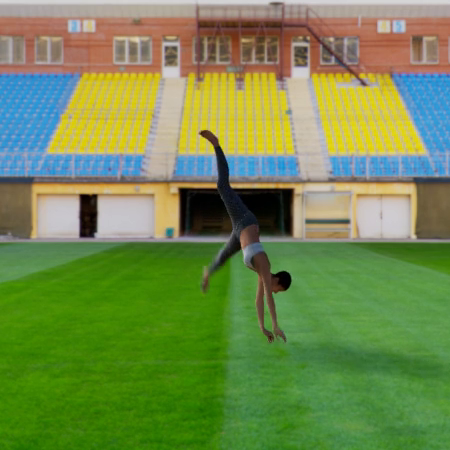} 
\vspace*{-0.2em}
\\

\raisebox{2em}{\textbf{Hispanic}}
& \includegraphics[width=0.14\linewidth]{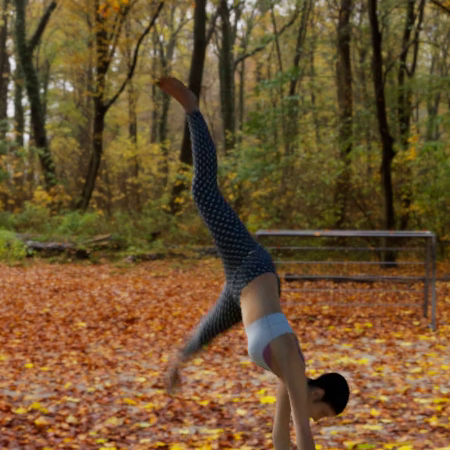}
& \includegraphics[width=0.14\linewidth]{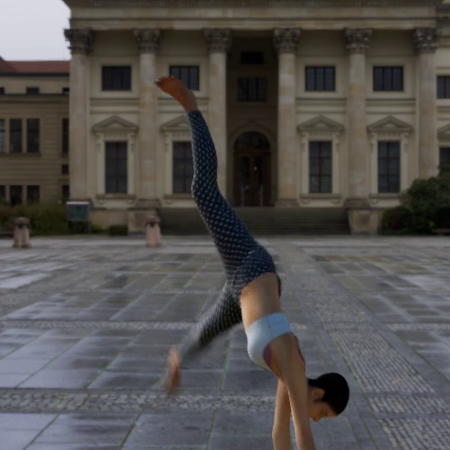}
& \includegraphics[width=0.14\linewidth]{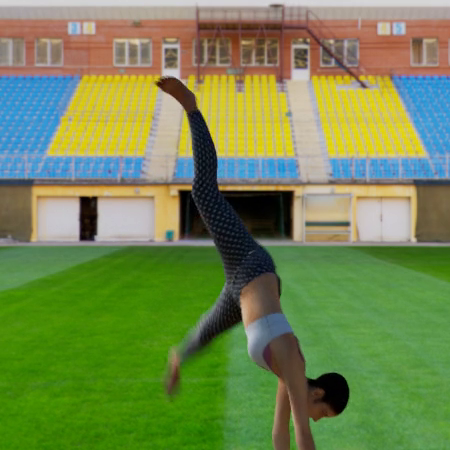}
& \includegraphics[width=0.14\linewidth]{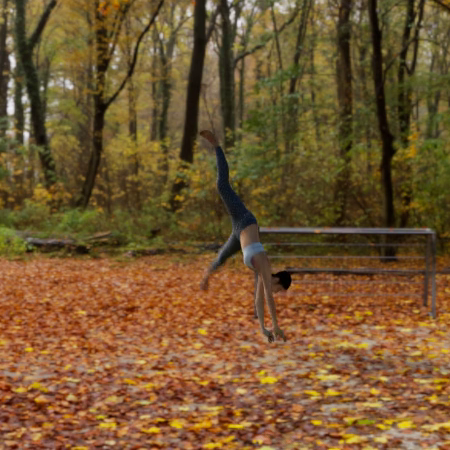}
& \includegraphics[width=0.14\linewidth]{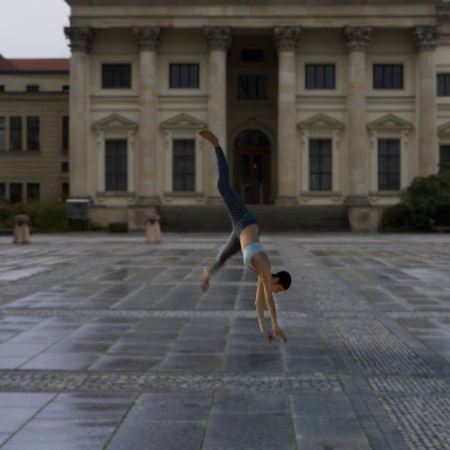}
& \includegraphics[width=0.14\linewidth]{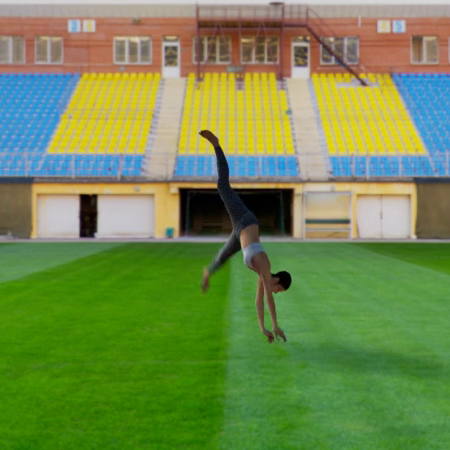} 
\vspace*{-0.2em}
\\

\raisebox{2em}{\textbf{Indian}}
& \includegraphics[width=0.14\linewidth]{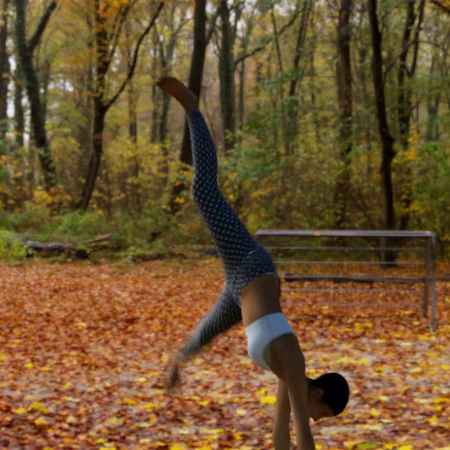}
& \includegraphics[width=0.14\linewidth]{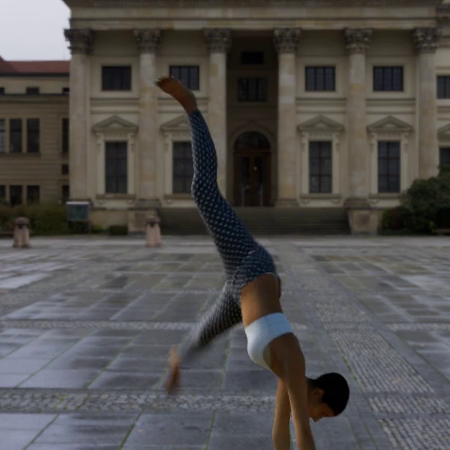}
& \includegraphics[width=0.14\linewidth]{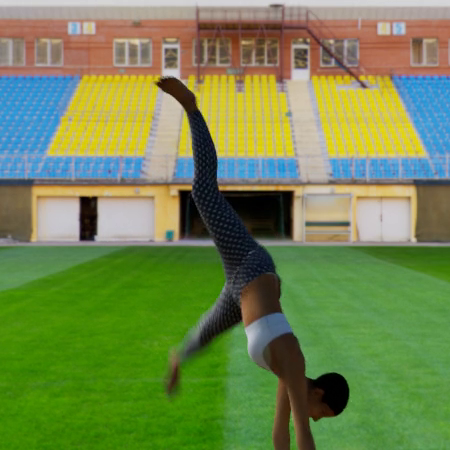}
& \includegraphics[width=0.14\linewidth]{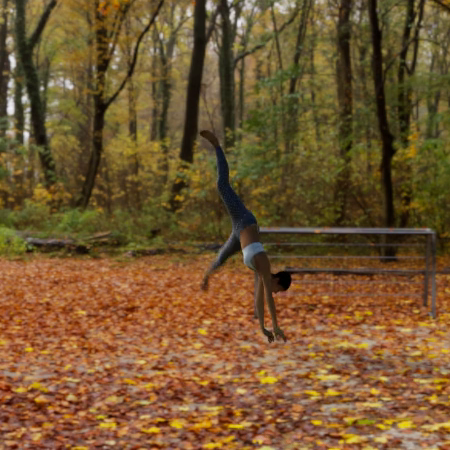}
& \includegraphics[width=0.14\linewidth]{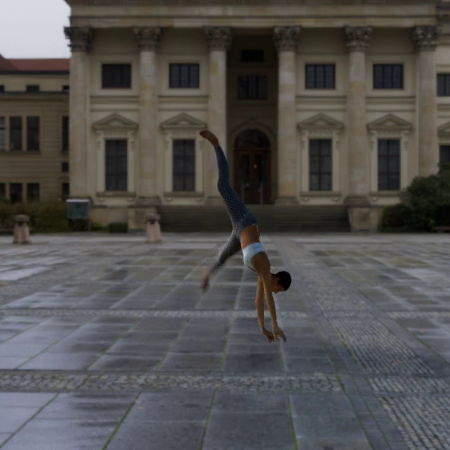}
& \includegraphics[width=0.14\linewidth]{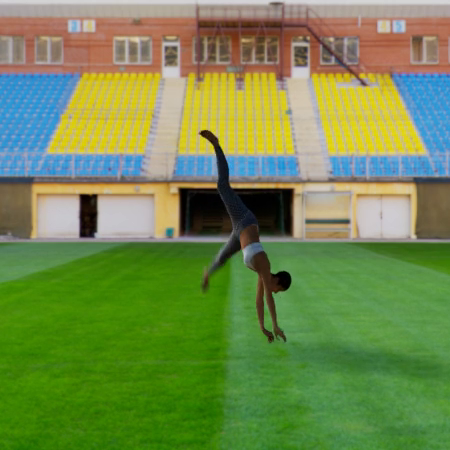} 
\vspace*{-0.2em}
\\

\raisebox{1.7em}{\textbf{\shortstack{Middle \\ Eastern}}}
& \includegraphics[width=0.14\linewidth]{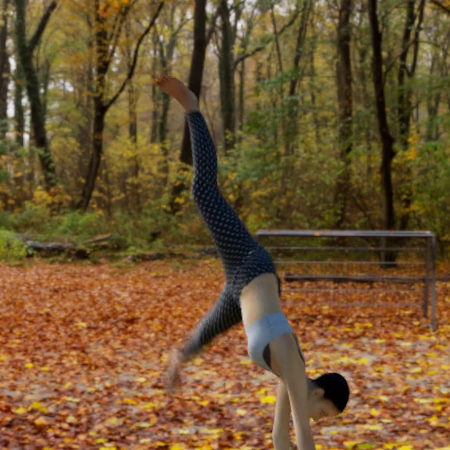}
& \includegraphics[width=0.14\linewidth]{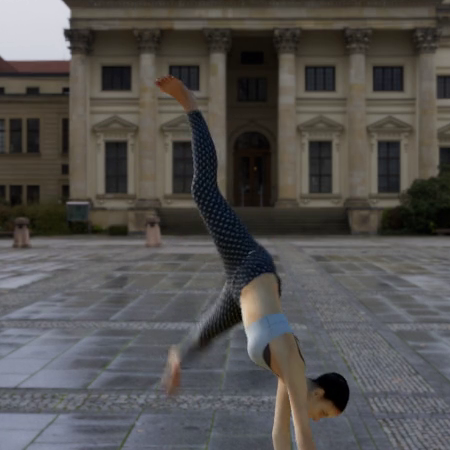}
& \includegraphics[width=0.14\linewidth]{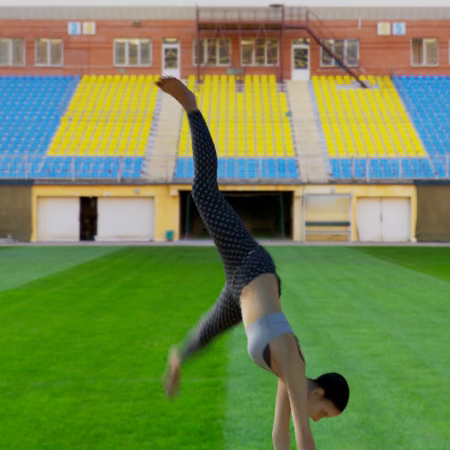}
& \includegraphics[width=0.14\linewidth]{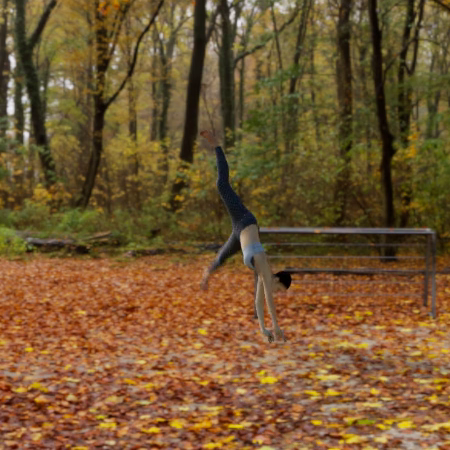}
& \includegraphics[width=0.14\linewidth]{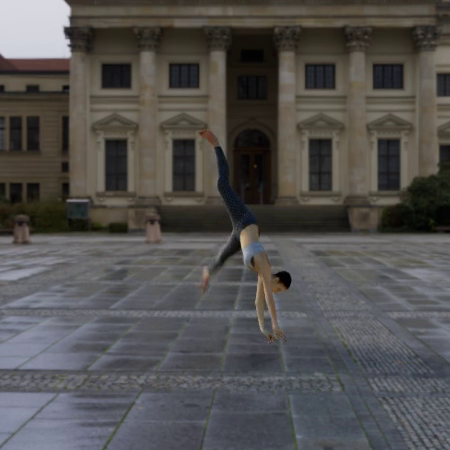}
& \includegraphics[width=0.14\linewidth]{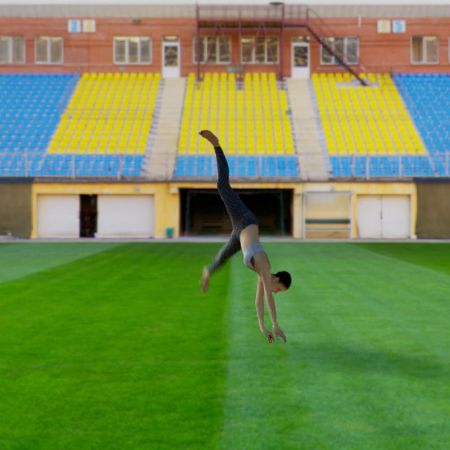} 
\vspace*{-0.2em}
\\

\raisebox{1em}{\textbf{\shortstack{South \\ East \\ Asian}}}
& \includegraphics[width=0.14\linewidth]{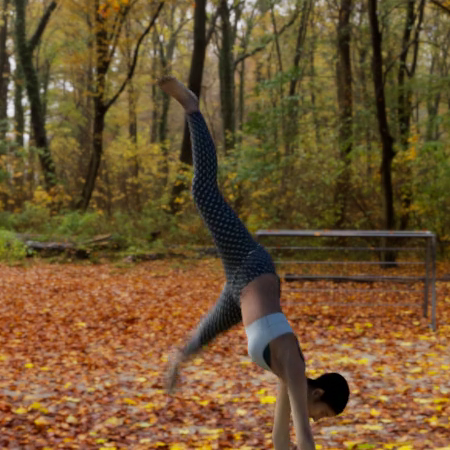}
& \includegraphics[width=0.14\linewidth]{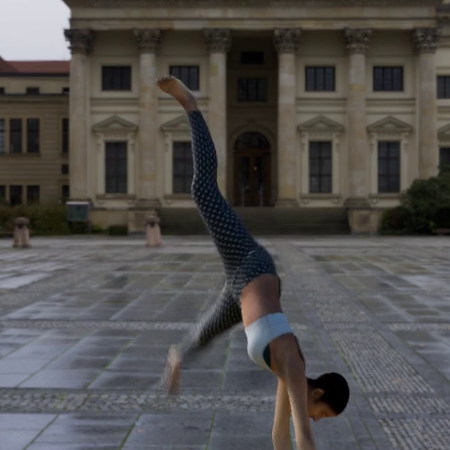}
& \includegraphics[width=0.14\linewidth]{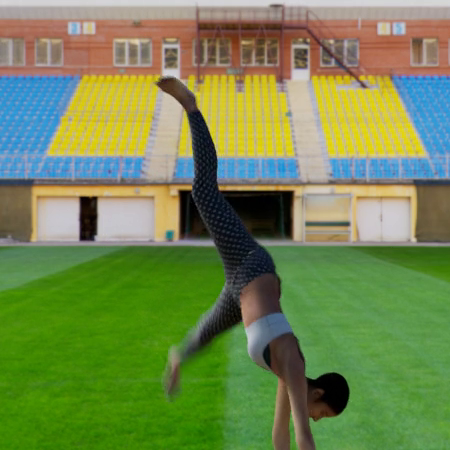}
& \includegraphics[width=0.14\linewidth]{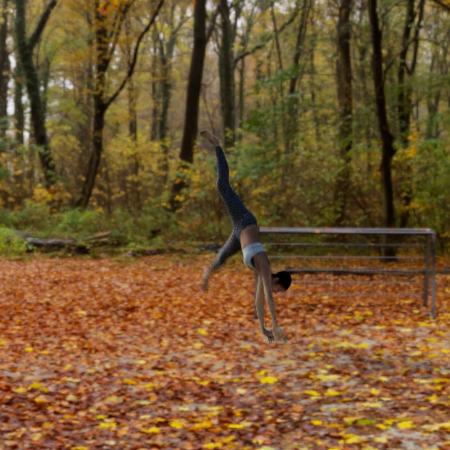}
& \includegraphics[width=0.14\linewidth]{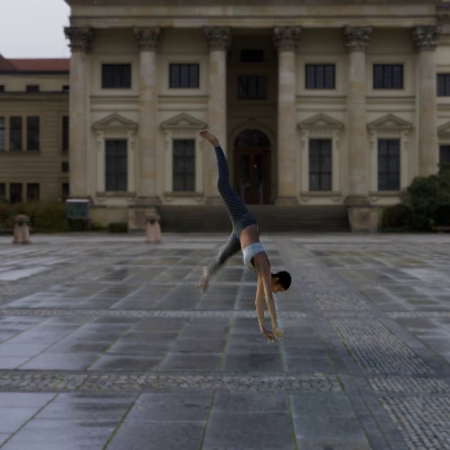}
& \includegraphics[width=0.14\linewidth]{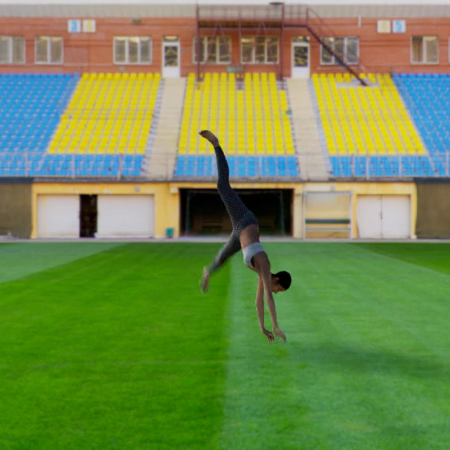} 
\vspace*{-0.5em}
\\

\bottomrule
\end{tabular}

\caption{
One motion of the cartwheel action, we see the same frame over all the different settings. This shows there is only the controlled difference we introduce in the synthetic data across individual attribute changes. 
}
\label{fig:cartwheel_motion_skin_viewpoints_backgrounds}

\end{figure}

\subsection{Our synthetic dataset: Ctrl-A-Bias}

To systematically evaluate model robustness to appearance-related biases in action recognition, we constructed the Ctrl-A-Bias synthetic dataset using the BEDLAM \cite{bedlam} framework to generate motions in the SMPL-X \cite{SMPL2015, SMPL_X2019} body model format, and rendered the videos in Unreal Engine. 
We used Python scripts to partially automate the pipeline and generate rendered video sequences, and the code is publicly available\footnotemark\footnotetext{Code: \url{https://github.com/ana-baltaretu/bias-action-recognition}}. 
Each video in the dataset corresponds to a single row in the accompanying CSV file, which includes metadata for reproducibility. 
Ctrl-A-Bias contains 8,400 short videos, created by independently varying five key dimensions:

\begin{itemize}[left=0pt]
    \item \textit{Skin color}: 7 texture categories from Meshcapade \cite{meshcapade}.
    \item \textit{Action category}: 20 human actions matched between labels of BEDLAM and Kinetics-400 \cite{kinetics}, see \autoref{subsec:dataset_generation_pipeline}.
    \item \textit{Motion variant}: 10 motion clips per action, from the ones available from BEDLAM \cite{bedlam}. Varied clothing texture when too few motions were available per action type.
    \item \textit{Camera viewpoints}: 2 fixed camera angles (near and far).
    \item \textit{Background}: 3 HDRI images from Poly Haven\footnotemark\footnotetext{Poly Haven \url{https://polyhaven.com/hdris}}.
\end{itemize}

This design ensures a dataset where each combination appears exactly once. 
The videos labeled ``initial'' are used to determine representative camera angles and background variations for the five HAR models described in \autoref{subsec:evaluated_har_models}. 
Based on this analysis, we select the best-performing viewpoint and background, and use the corresponding filtered subset of the dataset for bias evaluation. 
Our results show that model accuracy alone does not capture the full picture.

\subsection{Dataset generation pipeline}
\label{subsec:dataset_generation_pipeline}
Action-recognition models inherit the biases of the data on which they are trained. Public video datasets such as Kinetics-400 offer remarkable scale, but they supply little control over sensitive visual attributes like skin tone. To probe and understand these biases, we propose a synthetic‐dataset generation pipeline. 

\begin{enumerate}[left=0pt]
    \item We match labels between Kinetics \cite{kinetics} and BEDLAM \cite{bedlam} semantically, using SBERT \cite{sbert}.
    \item Using the list of most semantically matching labels, we randomly select multiple motions per action label, along with a random body type, which has associated clothing and clothing textures.
    \item We select one skin texture as part of the ``initial'' dataset, and we ran ablation studies to measure the influence of background and viewpoint on the model's accuracy for that action label.
    \item From these, we select the ``best'' background and viewpoint per action label, and we apply the remaining 6 skin textures to the animation, resulting in a total 7 videos where the actor performs the exact same motion, as seen in \autoref{fig:cartwheel_motion_skin_viewpoints_backgrounds}. 
    \item We compare the model accuracy across videos with the same motion where the skin color was altered. For an unbiased model, we expect the change in skin color to not affect the output labels.
\end{enumerate}

\section{Results}
\label{sec:Results}

Since we do not fine-tune the models, it is essential that our synthetic data closely resembles the distributions found in the datasets the models were originally trained on. 
To generate realistic and representative synthetic data, we made several design choices, focusing on factors that have been shown to significantly affect model performance \cite{kong2022human, wu2022survey, sun2022human, beddiar2020vision}: camera viewpoint and background environment, see \autoref{subsec:key_design_principles}. 
We conducted ablation studies to assess how changes in camera position and background influence accuracy.
Crucially, if a model fails to understand the scene, any fairness analysis would be irrelevant, because there would be too much random variation. 
Therefore we only included models that met a minimum performance threshold. 
Lastly, we examined how sensitive each selected model is to changes in skin color and tested whether the differences in predictions were statistically significant across skin tones.

\begin{figure}[ht]
\centering
\includegraphics[width=\linewidth]{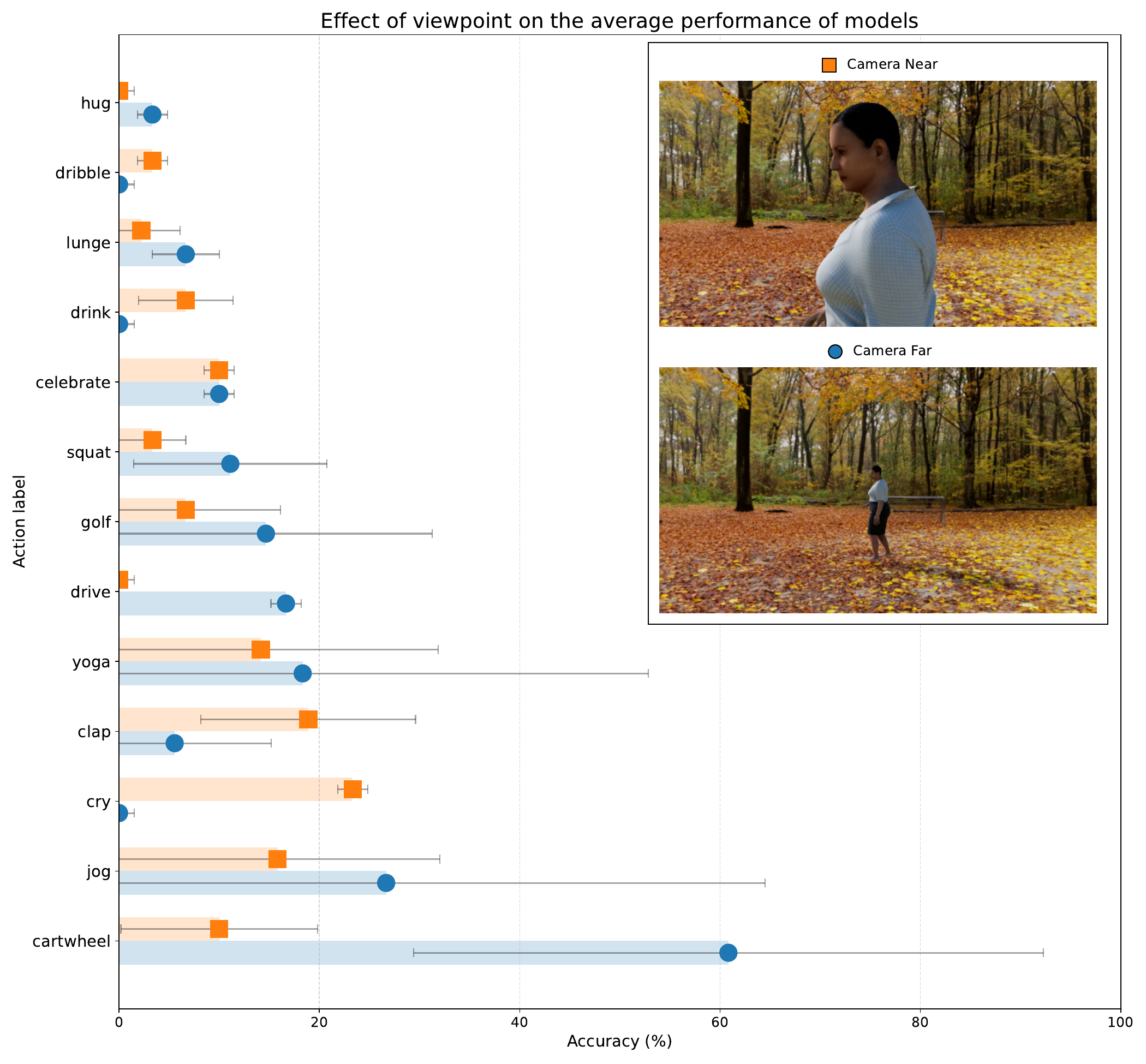}
\caption{Impact of Viewpoint on action recognition accuracy. Mean accuracy and standard deviation over models for each action class is shown from ``Near'' and ``Far'' camera positions. Complex, full-body actions (cartwheel, jog, yoga) lose accuracy at the Near viewpoint, while simple, localized actions (drink, cry) show the opposite trend. This demonstrates that camera placement can significantly impact model evaluation especially for complex actions and highlights the necessity of using representative viewpoints when generating synthetic data for testing.}
\label{fig:camera_distance_ablation}
\end{figure}

\subsection{Minimizing confounding factors}
To keep the scale of our synthetic benchmark manageable, we fixed most scene parameters. 
Every clip shows a single actor who begins the motion at screen center under uniform Unreal-Engine lighting. 
For factors without an obvious default (viewpoint and background) we first ran ablation studies, then picked the setting that yielded the most stable model performance.

\textbf{Camera position variation.} 
To isolate viewpoint sensitivity we rendered every motion from two viewpoints: Near (around 1m from the actor, waist-up framing) and Far (around 6m away, full-body framing). 
These settings mirror tight and wide shots, allowing direct comparison to real datasets. 
As can be seen in \autoref{fig:camera_distance_ablation}, changing the camera viewpoint can shift top-1 accuracy significantly, and the direction of the effect is action-specific. 
Full-body activities suffer when viewed up close, because limbs leave the frame and temporal cues are lost, whereas compact upper-body actions benefit from a closer viewpoint.


\begin{figure}[ht]
\centering
\includegraphics[width=\linewidth]{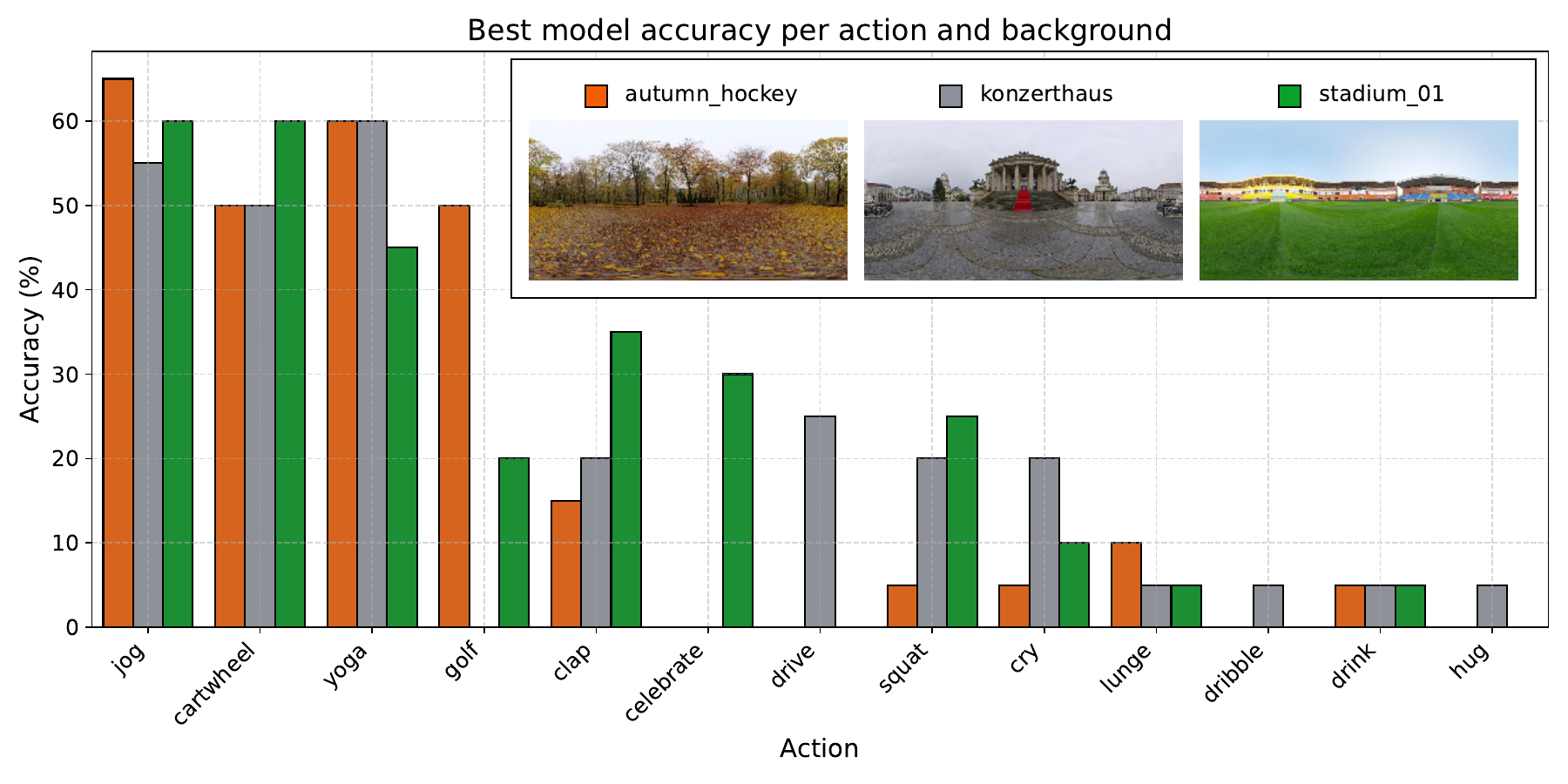}
\caption{Impact of Background on the accuracy of models to recognize certain actions. The height of the bars represents the highest prediction accuracy of a background for any of the 5 evaluated models for each action label. We can see from this graph that some actions like jog and cartwheel have similar accuracies no matter the background, while accuracies of  other actions like golf or celebrate are significantly impacted if the background differs from the training data.}
\label{fig:background_ablation}
\end{figure}

\textbf{Background variation.} 
To evaluate the influence of background on model predictions, we rendered each action across three outdoor scenes: an autumn park, a grey urban plaza, and a bright stadium, each with a distinct color palette. As backgrounds occupy most of the frame, we expected them to act as strong visual cues. 
The results in \autoref{fig:background_ablation} confirm this: full-body motions like jog and cartwheel remain accurate across settings, while context-driven actions like golf or celebrate vary significantly by background. 
Some scenes, like the autumn park, consistently biased models toward a small set of actions (e.g., golf), regardless of the actual motion. 
To control for this bias, we fixed each action to the background with the most reliable accuracy before varying other factors.

These findings confirm that non-motion visual cues such as viewpoint and background environment affect model performance and must be controlled when probing for biases.
\newpage
\begin{figure}[htp]
    \centering
    \includegraphics[width=0.95\linewidth]{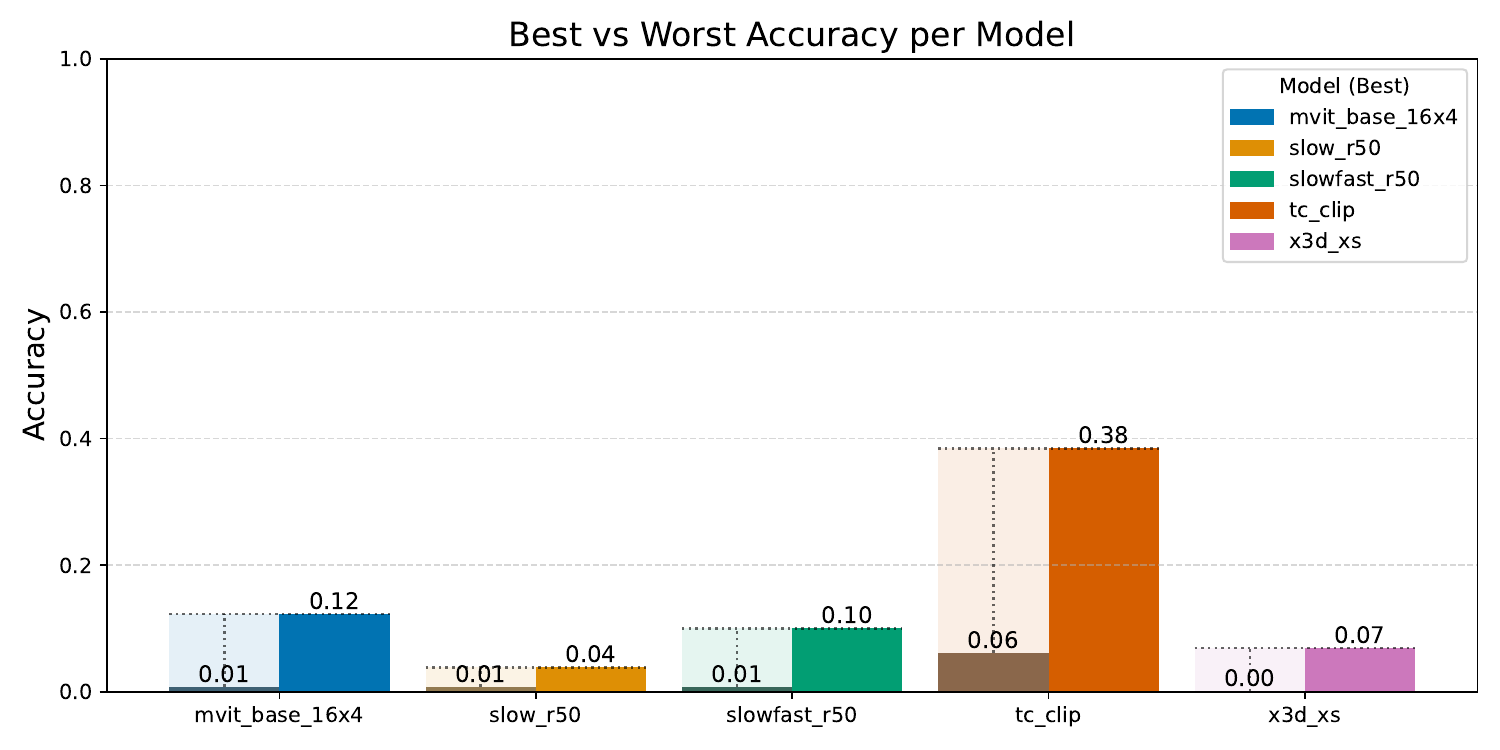}
    \caption{Model performance on the baseline synthetic dataset. For each model, the left bar shows the average accuracy under suboptimal viewpoint and background settings, while the right bar shows the accuracy with the best-performing configuration. The results highlight that selecting more suitable settings per action label improves accuracy for all models, making our baseline more comparable to real datasets.}
    \label{fig:best_vs_worst_joint_barplot_combined}
\end{figure}
\subsection{How do models generalize to synthetic actions?}
\label{subs:baseline}

Before evaluating fairness, we first validated whether each model could reliably interpret our synthetic setup. 
This ensures that any observed biases are not simply due to models not understanding the actions in the first place. 
For this, we selected one viewpoint and one background per action label, the ones that yielded the best average accuracy across models in the ablation studies.
\autoref{fig:best_vs_worst_joint_barplot_combined} shows the performance of each model under the ``best'' and ``worst'' case attribute combinations. 
All models improve with more favorable conditions, confirming that our synthetic design allows models to generalize to some extent. 
However, certain models such as \texttt{Slow\_R50} and \texttt{X3D\_XS} still perform poorly, even under improved settings. 
Their near-random accuracy suggests they lack a meaningful understanding of the actions, making any bias evaluation unreliable.
To ensure a fair comparison in later experiments, we only continue evaluation on models \texttt{MViT}, \texttt{SlowFast}, and \texttt{TC-Clip}, which demonstrated better accuracy than the others and showed consistent predictions on the synthetic data.

\begin{figure}[htp]
    \centering
    \includegraphics[width=\linewidth]{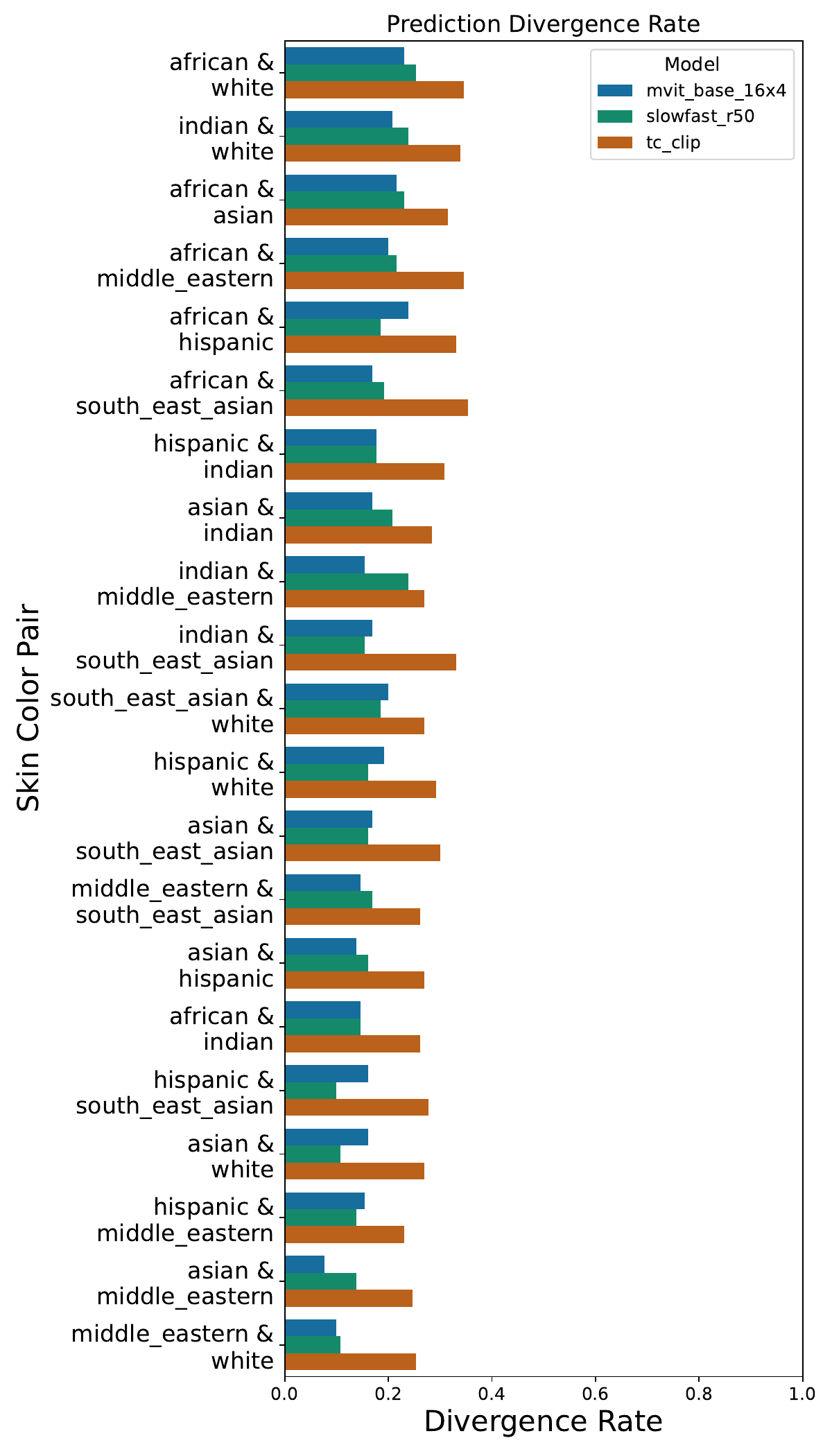}
    \caption{Proportion of action label predictions that differ when an action is performed in the same setting by actors with different skin tones. A lower divergence rate suggests less reliance on visual cues like skin color, an ideal trait when the task depends on motion rather than appearance. Notably, TC-CLIP consistently exhibits higher divergence rates across all pairs, indicating greater sensitivity to skin tone despite achieving higher average accuracy on the baseline synthetic dataset. This suggests that improved accuracy does not necessarily imply improved fairness.}
    \label{fig:model_divergence_rate_per_skincolor_pair}
\end{figure}

\subsection{How does skin tone affect predictions?}

To assess the extent to which models rely on visual appearance (specifically skin color), rather than relying motion, we compute the prediction divergence rate between pairs of skin colors ($s_1$, $s_2$). 
For a given pair of skin textures, we define the divergence as the number of video instances for which the predicted action label differs when the same motion is performed by actors of skin color $s_1$ vs $s_2$, and we calculate divergence rate formally as:


\begin{equation}
    \label{eq:lbs}
    \text{Divergence Rate}(s_1, s_2) = \frac{\sum_{i=1}^{N}  y^{s_1}_i \ne y^{s_2}_i}{N}
\end{equation}

\noindent Where:
\begin{itemize}
    \item $y$ is the predicted label for a video 
    \item ($s_1$, $s_2$) are paired videos showing the same motion $i$ with different skin colors
    \item $N$ is the total number of motions
\end{itemize}

\noindent This metric assesses how frequently the model changes its prediction solely due to a change in skin tone, as motion (not appearance) should be the primary cue for classification.
\autoref{fig:model_divergence_rate_per_skincolor_pair} shows the divergence rates across various skin color pairs for three models. 
Ideally, a model that is invariant to appearance would exhibit a divergence rate close to zero for all pairs.
However, we observe that TC-CLIP, despite achieving the highest overall accuracy on the synthetic dataset (\autoref{fig:best_vs_worst_joint_barplot_combined}), consistently demonstrates the highest divergence rates across all skin color pairs. 
This suggests that its predictions are more sensitive to variations in skin tone than the other two models.
These results highlight that accuracy may come at the cost of fairness, especially if it is partly driven by reliance on appearance-based cues.

While all models show some variability across skin tone pairs, a truly biased model would demonstrate significantly higher divergence between a particular skin color combinations compared to the other pairs, because that means it consistently relies on the skin tone to predict actions. 
These outliers would point to model confusion being driven disproportionately by specific appearance factors, which we investigate further in the next section.





\subsection{Do any skin tone pairs cause significantly more prediction changes?}
To investigate whether models are disproportionately affected by specific skin color modifications, we statistically compare divergence rates between all skin tone pairs, shown in \autoref{fig:pairwise_significance}. 
As noted earlier, a biased model would show significantly higher prediction changes for certain skin color pair, pointing to a reliance on visual cues over motion.
In the raw p-values (top row), we observe several pairs that reach significance thresholds, suggesting bias in models. 
However, once Bonferroni correction \cite{armstrong2014use} is applied to control for multiple comparisons (bottom row), nearly all significant results disappear.
This outcome suggests that, while some skin tone changes may lead to more prediction changes than others, the evidence is not strong enough to confirm consistent bias toward particular skin colors.
One possible explanation is that the models are generally appearance-sensitive but not selectively biased. 
Another explanation is that the dataset shows limited realism, not enough to expose more systematic patterns of bias.

\begin{strip}
\centering

\setlength{\tabcolsep}{1em} 
\renewcommand{\arraystretch}{0.8} 
\centering

\begin{tabular}{c c c c}
& \textbf{SlowFast} & \textbf{MViT} & \textbf{TC-Clip} \\

\rotatebox{90}{\makebox[10em][c]{\textbf{Raw}}} &
\includegraphics[width=0.26\linewidth]{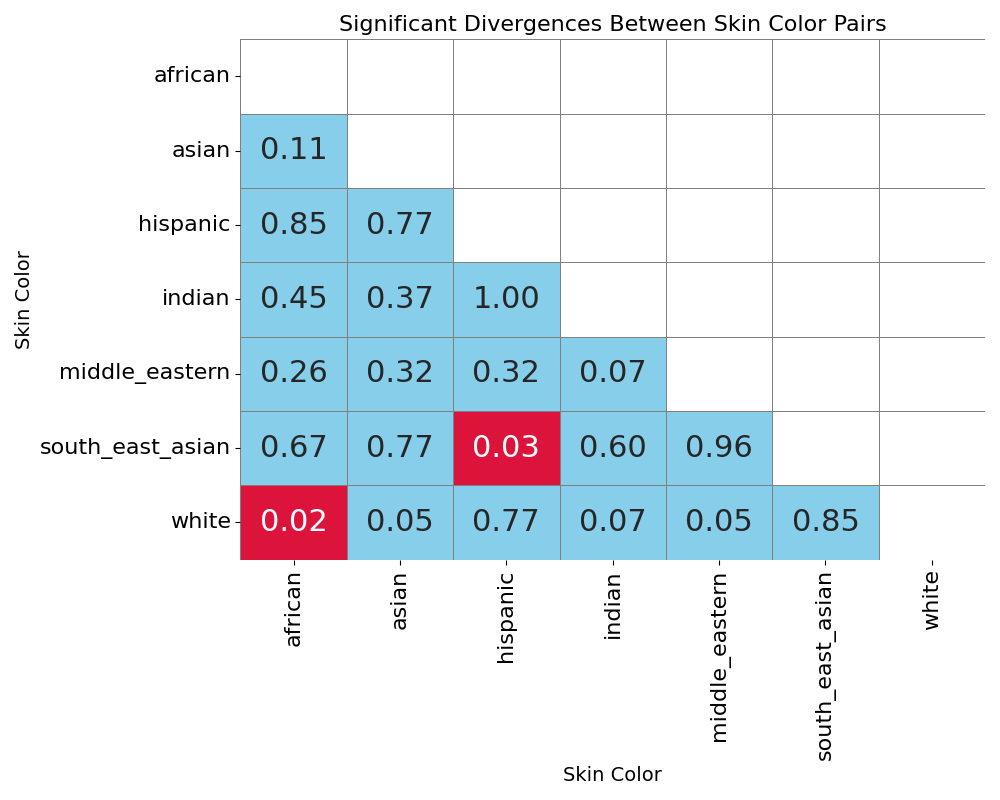} &
\includegraphics[width=0.26\linewidth]{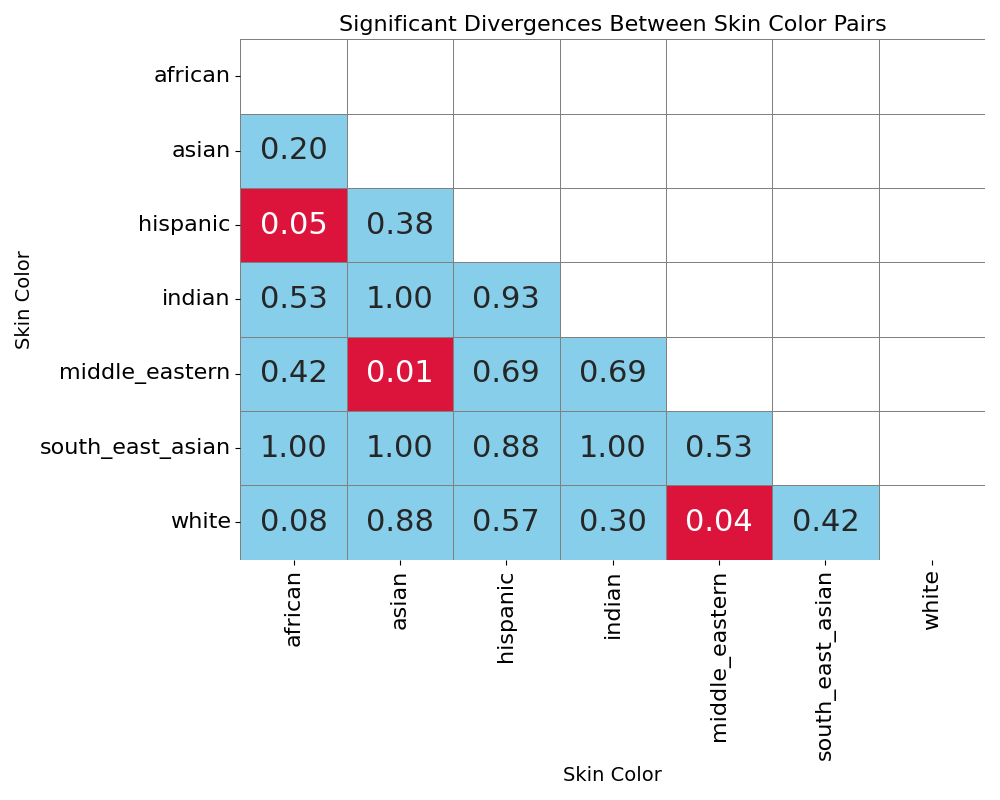} &
\includegraphics[width=0.26\linewidth]{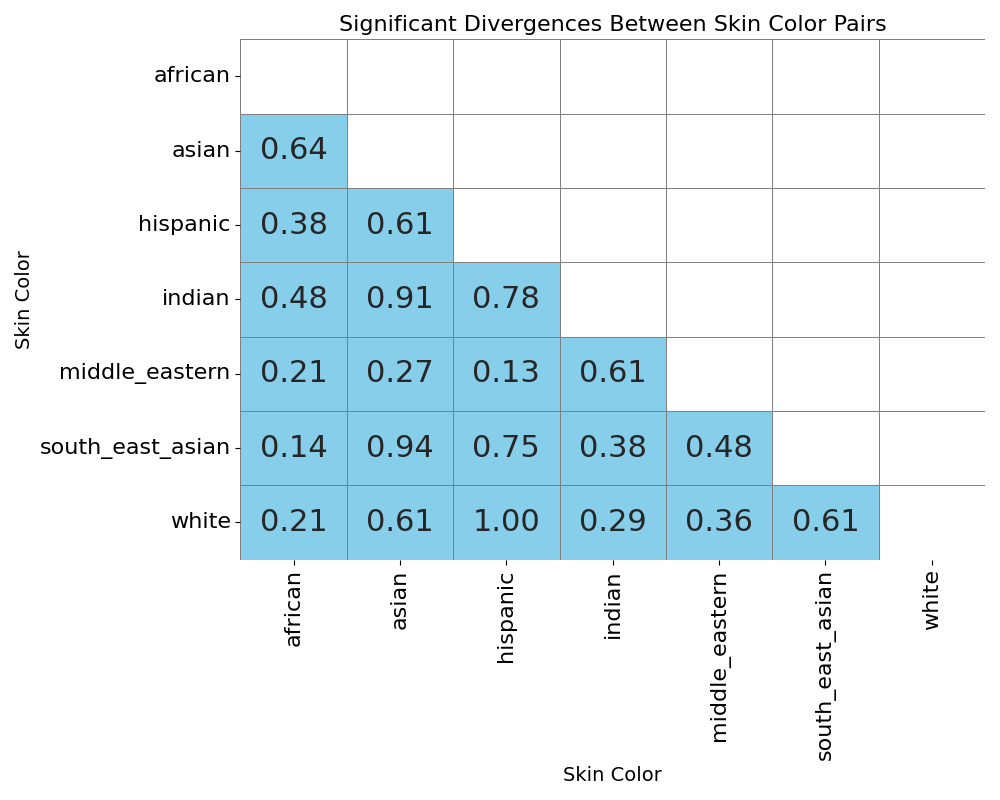} \\

\rotatebox{90}{\makebox[10em][c]{\textbf{Adjusted}}} &
\includegraphics[width=0.26\linewidth]{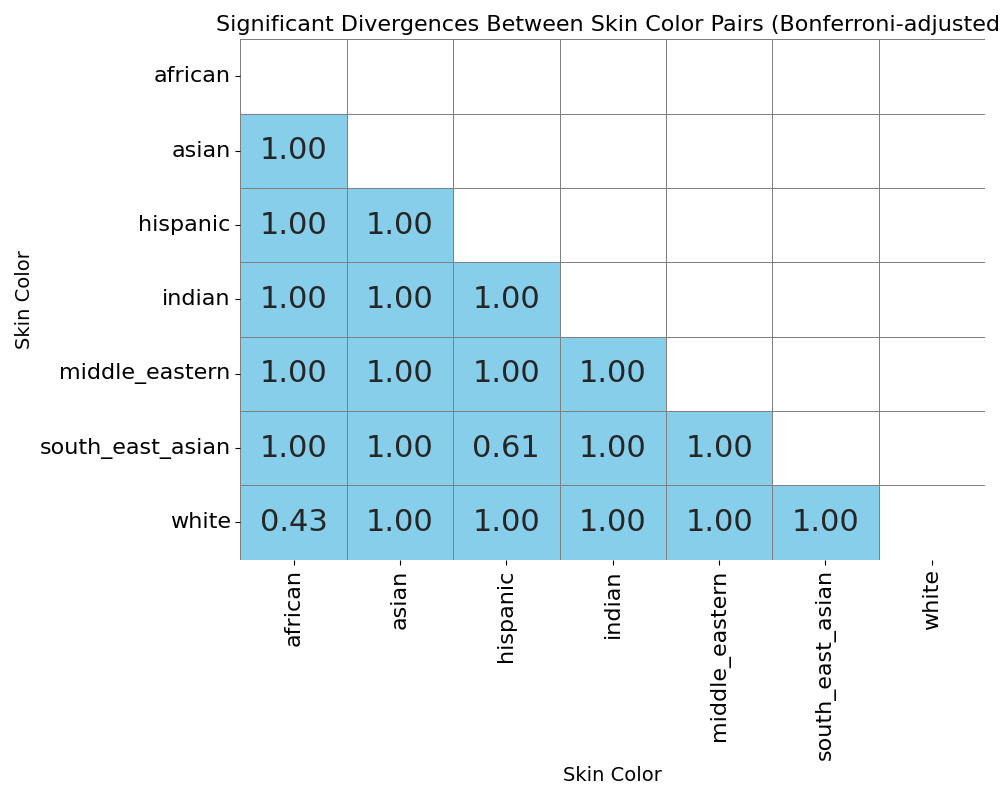} &
\includegraphics[width=0.26\linewidth]{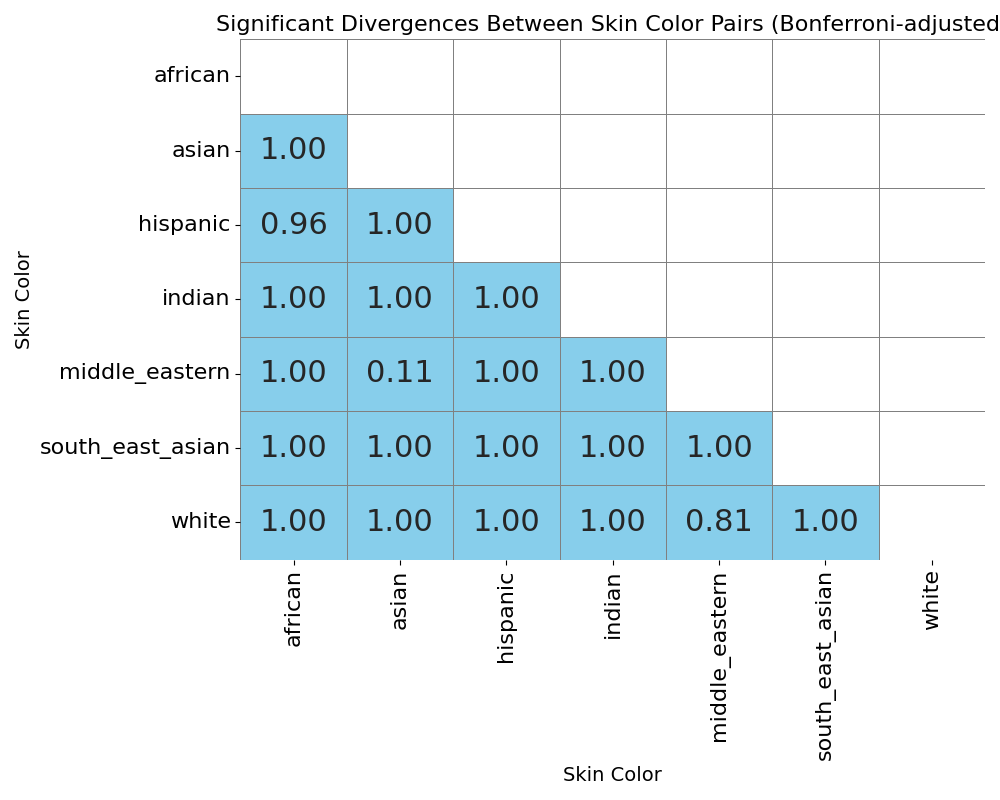} &
\includegraphics[width=0.26\linewidth]{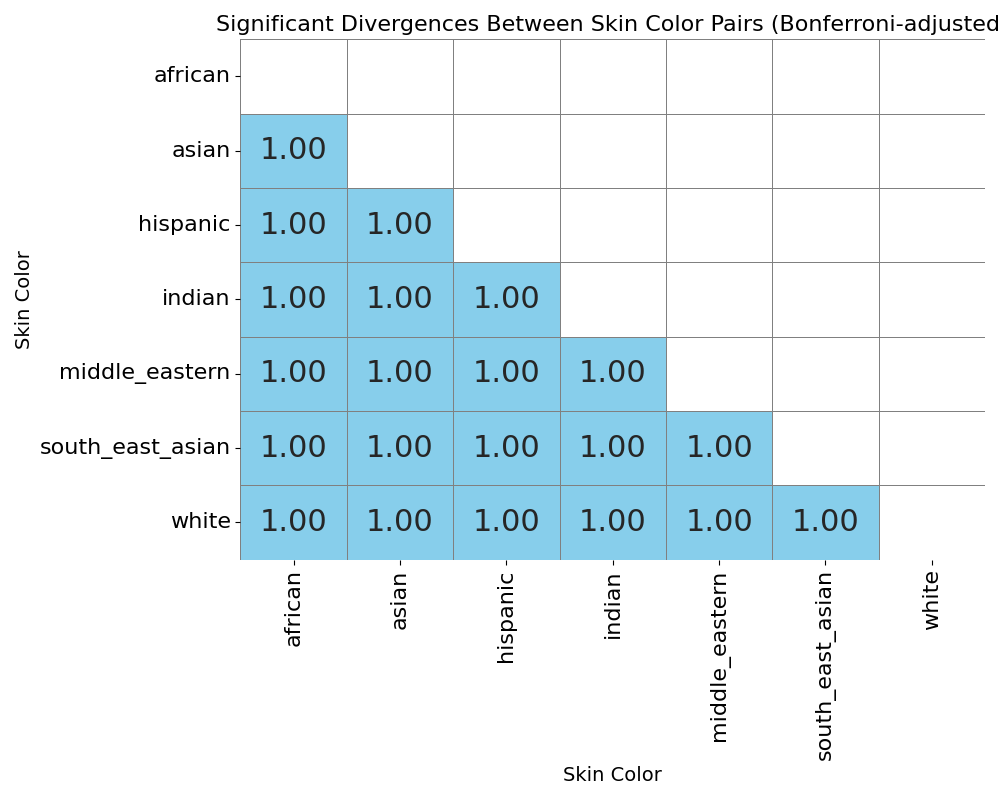} \\
\end{tabular}

\captionof{figure}{
Statistical significance of prediction divergence between skin color pairs.
Top row: raw p-values for each model and skin color pairs. 
Red cells indicate that prediction changes are significant ($p < 0.05$).
Bottom row: Bonferroni-adjusted p-values correcting for multiple comparisons. 
After correction, most significant effects disappear.
This may suggest that the models are not systematically biased toward specific skin color pairs, though it is also possible that such bias was not detected due to limitations in data realism.
}
\label{fig:pairwise_significance}
\end{strip}

\vspace*{-1cm}

\section{Discussion}
\label{sec:discussion}
We reflect on key ethical and practical limitations of our approach and outline future directions, including issues of identity representation, potential misuse, and dataset scope.

\subsection{Ethical considerations}
This work involves sensitive aspects of identity and fairness in action recognition. 
We reflect on the ethics of using synthetic human models for auditing, focusing on harmful labels, potential misuse, and the intent of our analysis.

\textbf{Demographic labels.}
Although our dataset includes labels such as ``female'' or ``white'' we acknowledge that these are simplifications that may reinforce binary or essentialist understandings of gender, race, or identity, which we do not support. 
Ideally, demographic variation would be modeled along continuous, multidimensional axes. 
Unfortunately, the SMPL \cite{SMPL2015, SMPL_X2019} framework offers only a limited set of discrete body shapes: male, female, and a neutral based on biological sex, and skin textures with categorical labels. 
We use these constraints to approximate demographic variation and show potential biases in model behavior, but we want to the reader to be aware of their interpretive limitations.

\textbf{Risk of misuse and clarification of intended purpose.}
We recognize that tools for bias auditing can be misused in harmful ways in downstream applications, such as ranking individuals by demographics in surveillance or advertising.
We do not support such uses. 
Our dataset is not for profiling, rather our goal is to enable transparent testing of models to uncover disparities in performance across demographic attributes.

While some people may view measuring bias without proposing mitigation strategies as ethically shallow, we see this work as a foundational step. Our goal is to surface and quantify disparities in model behavior, laying the groundwork for future research to develop and explore corrective measures.

\subsection{Limitations and Future work}
Our project enables structured auditing of action recognition models with synthetic data, but is limited by the scale of ablations and motion diversity. 
We outline these constraints and suggest future improvements.

\textbf{More ablations.}
We focused on background and viewpoint variation, but other factors like lighting, action speed, actor position, and number of people may also affect model performance (\autoref{subsec:HAR_so_challenging}). 
These dimensions could be explored through similar ablation studies. 
However, we were limited to two ablations due to the significant time required to generate the data. 
Specifically, we produced six batches of 1400 videos each (one batch per background-viewpoint combination), with each batch taking 5–6 hours to generate. 
In addition to the generation time, inference on all five models added another 6 hours to the total generation time. 
Given these constraints, we were unable to scale up to more dimensions. 
Future work could explore additional ablations across the other relevant factors mentioned above, and we recommend testing each property independently, to not exponentially expand the dataset.

\textbf{Limited motions.}
Our current setup relies on baked animations from BEDLAM \cite{bedlam}, which are sourced from the AMASS dataset \cite{AMASS}. 
While this provides a broad set of motions, it remains limited to predefined actions and may not generalize well to specific application domains. 
One promising direction for future work is to leverage Meshcapade \cite{meshcapade}, which allows generating custom motion sequences through its motion-to-text feature tailored to specific tasks or datasets. 
This could enable the creation of more targeted and varied ablations, especially when aligning synthetic motions with the action categories present in downstream benchmarks.
We did not pursue this option in the current project for several reasons, one of them is that it lacks clothing realism and requires integration with BEDLAM’s clothing pipeline\footnotemark\footnotetext{BEDLAM clothing repository: \url{https://github.com/PerceivingSystems/bedlam_clothing}}, which was too complex for this project.
Still, it holds promise for generating diverse and detailed motions in future work.

\section{Conclusions}
\label{Conclusions}

We presented a framework for auditing Human Action Recognition (HAR) models using synthetic video data with controlled appearance variations. 
By isolating attributes like skin tone, we evaluated whether models rely on visual cues unrelated to motion, and some key findings include:

\begin{itemize}[left=0pt]
    \item Viewpoint and background strongly affect model accuracy.
    \item Only a subset of models generalized well to synthetic data.
    \item All models showed sensitivity to skin tone changes.
    \item No significant bias between specific skin tone pairs was found after correction.
\end{itemize}

Our results highlight that appearance can still influence predictions even in the absence of bias towards a certain group. 
We encourage researchers and developers to use synthetic interventions to evaluate their models, especially in light of regulations like the EU AI Act \cite{down2021proposal}. 
While we focused on Human Action Recognition, the presented methodology could also be applied to other video-based tasks such as video understanding or video captioning. 
We see this work as a first step: a foundation on which more domain-specific bias auditing tools and mitigation strategies can be built.

\newpage
\printbibliography 

\clearpage

\appendix
\section{Models comparison when changing between skin colors}

\begin{strip}
  \centering
  \includegraphics[width=\textwidth]{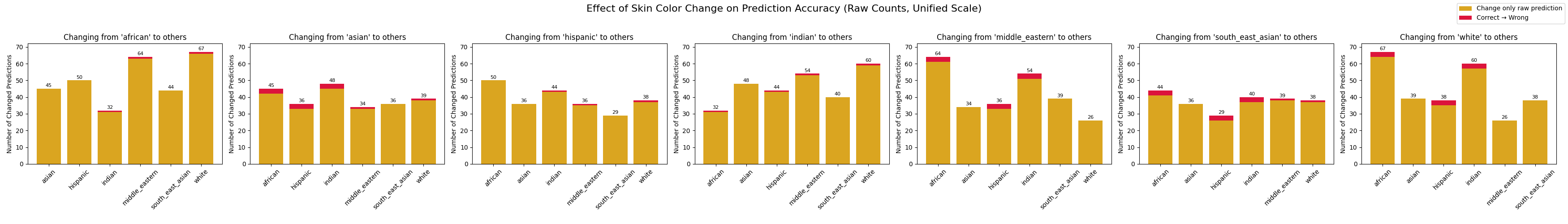}
  \captionof{figure}{Slowfast, differences when changing between skin colors.}
  \label{fig:subfig1}
  \vspace{1em}
  \includegraphics[width=\textwidth]{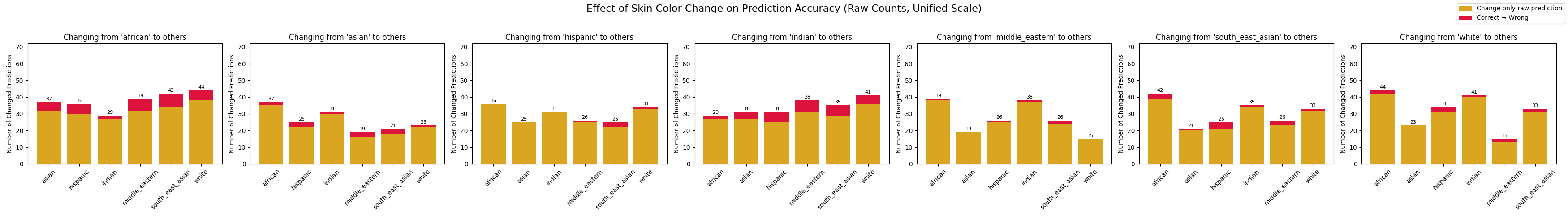}
  \captionof{figure}{Mvit, differences when changing between skin colors.}
  \label{fig:subfig2}

  \includegraphics[width=\textwidth]{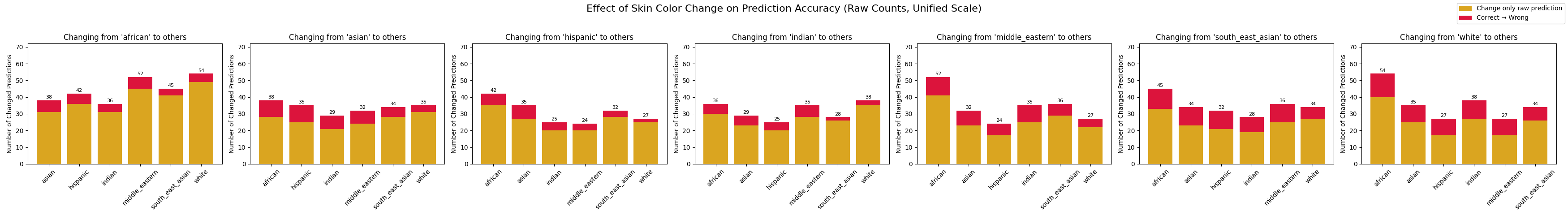}
  \captionof{figure}{TC-clip, differences when changing between skin colors.}
  \label{fig:subfig3}
\end{strip}

\autoref{fig:subfig1}, \autoref{fig:subfig2} and \autoref{fig:subfig3}  For \texttt{maxed\_out\_distributions\_on\_kinetics\_konzerthaus}

\clearpage

\begin{figure}[htp]
    \centering
    \includegraphics[width=0.9\linewidth]{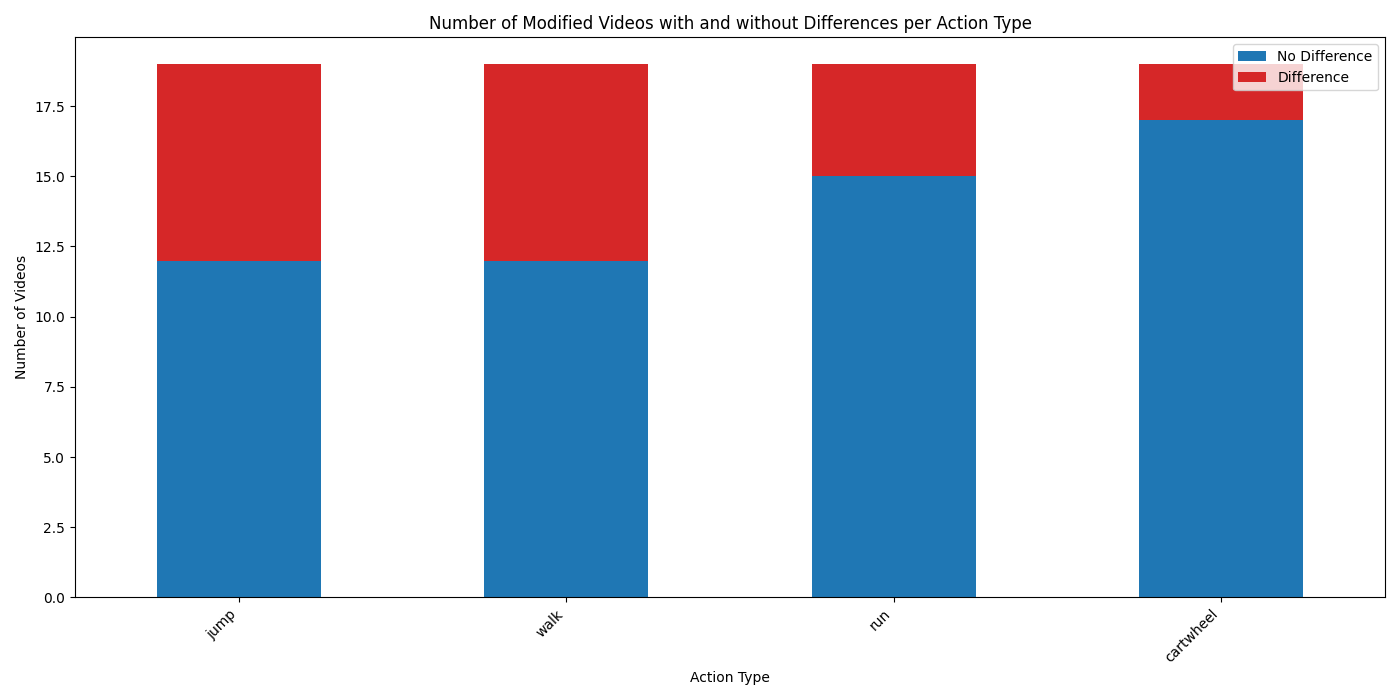}
    \caption{how many differences there are when changing to another skin color, out of all the modified videos per that action type for SlowFast model}
    \label{fig:modified_videos_differences_slowfast}
\end{figure}

\begin{figure}[htp]
    \centering
    \includegraphics[width=0.9\linewidth]{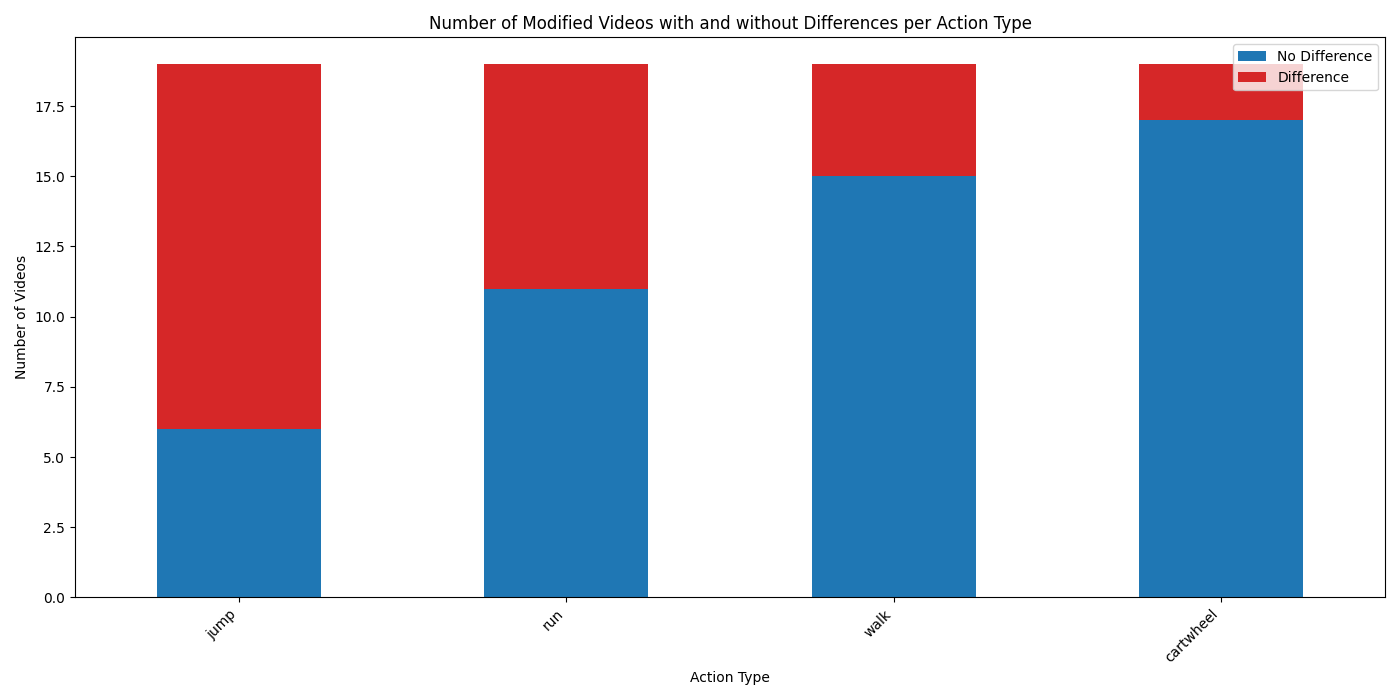}
    \caption{how many differences there are when changing to another skin color, out of all the modified videos per that action type for MVIT model}
    \label{fig:modified_videos_differences_mvit}
\end{figure}

\begin{figure}[htp]
    \centering
    \includegraphics[width=0.9\linewidth]{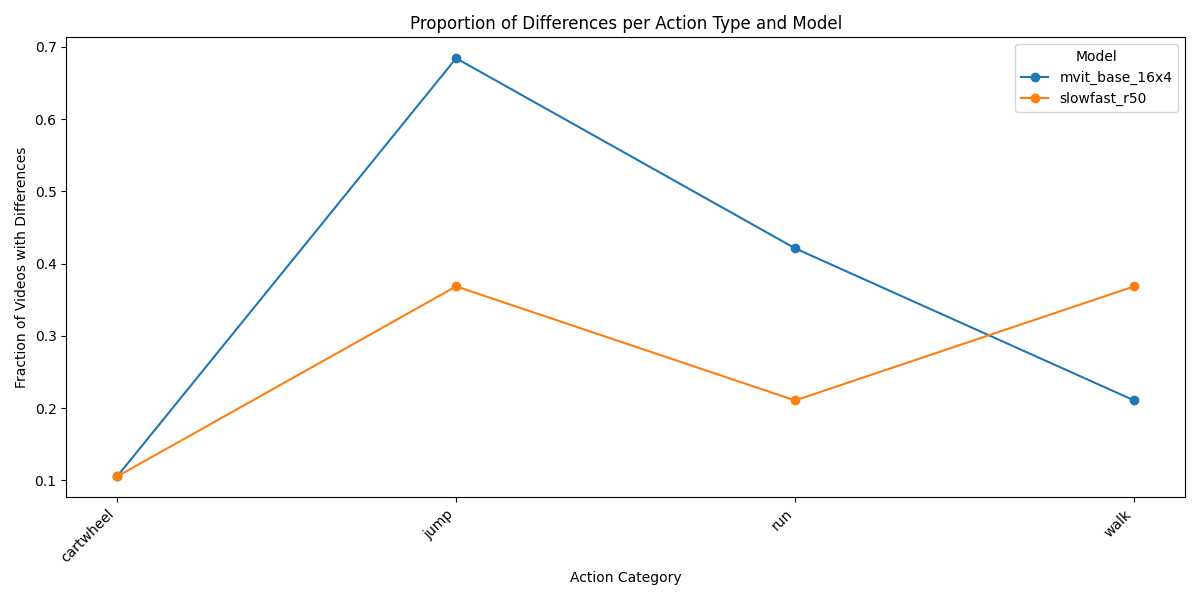}
    \caption{percentage differences there are out of all the modified videos per that action type for 2 models}
    \label{fig:modified_videos_differences_percentage}
\end{figure}

\autoref{fig:modified_videos_differences_slowfast},
\autoref{fig:modified_videos_differences_mvit} and
\autoref{fig:modified_videos_differences_percentage} are on old dataset with handpicked actions \texttt{cartwheel, jump, run, walk}

\begin{figure}[ht]
  \centering
  \begin{subfigure}[t]{0.49\linewidth}
    \centering
    \includegraphics[width=\linewidth]{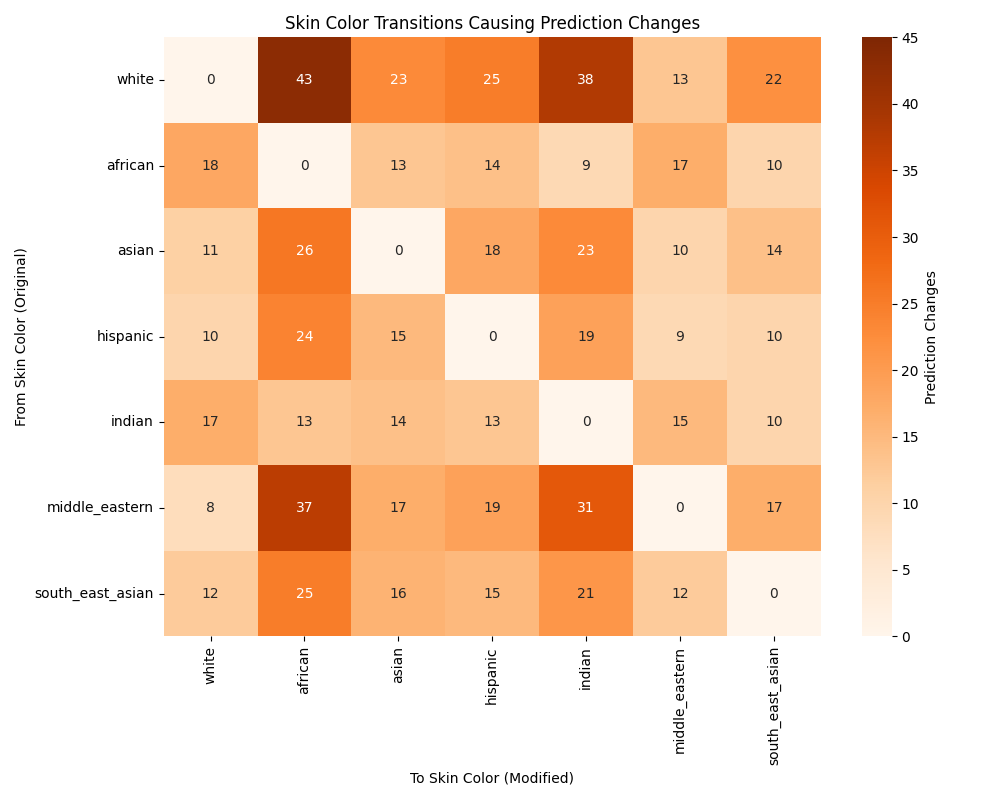}
    \subcaption{SlowFast differences}
    \label{fig:subfig1}
  \end{subfigure}
  \hfill
  \begin{subfigure}[t]{0.49\linewidth}
    \centering
    \includegraphics[width=\linewidth]{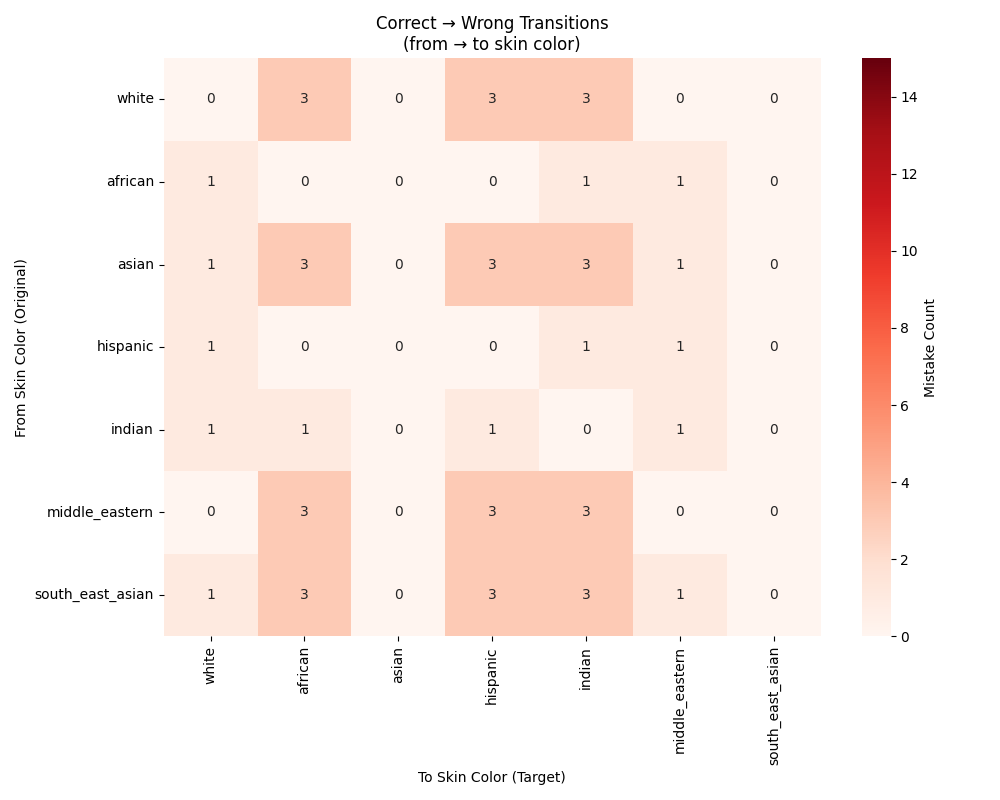}
    \subcaption{SlowFast errors}
    \label{fig:subfig1}
  \end{subfigure}
  \\
  \begin{subfigure}[t]{0.49\linewidth}
    \centering
    \includegraphics[width=\linewidth]{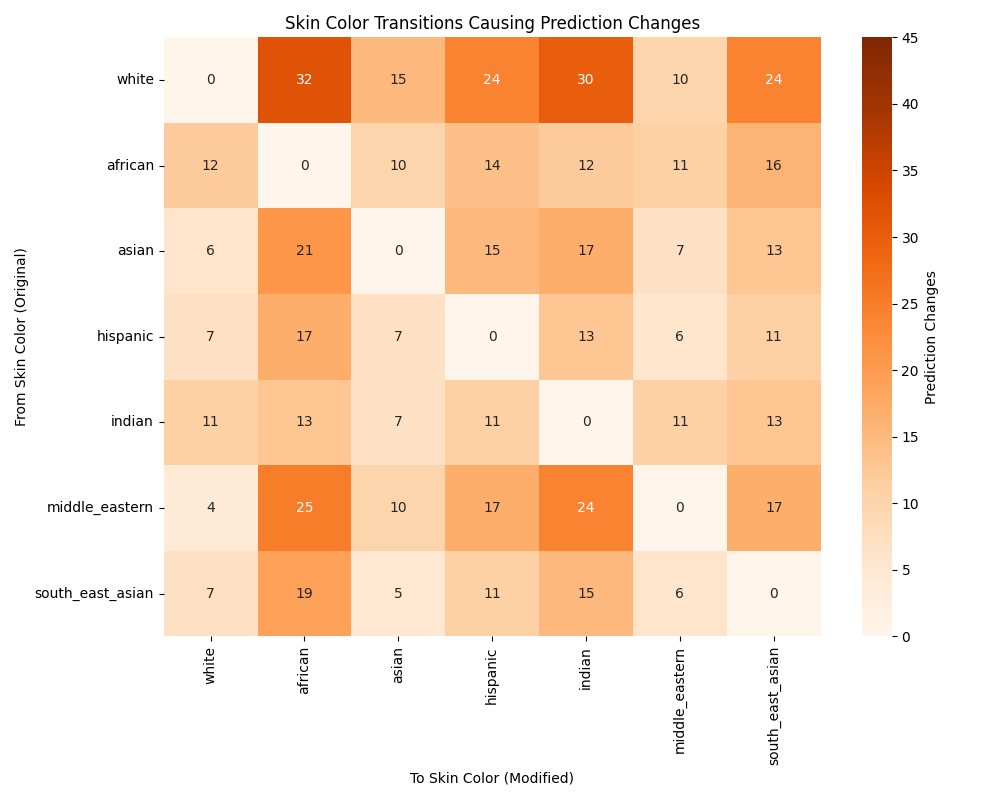}
    \caption{MViT differences}
    \label{fig:subfig2}
  \end{subfigure}
  \hfill
  \begin{subfigure}[t]{0.49\linewidth}
    \centering
    \includegraphics[width=\linewidth]{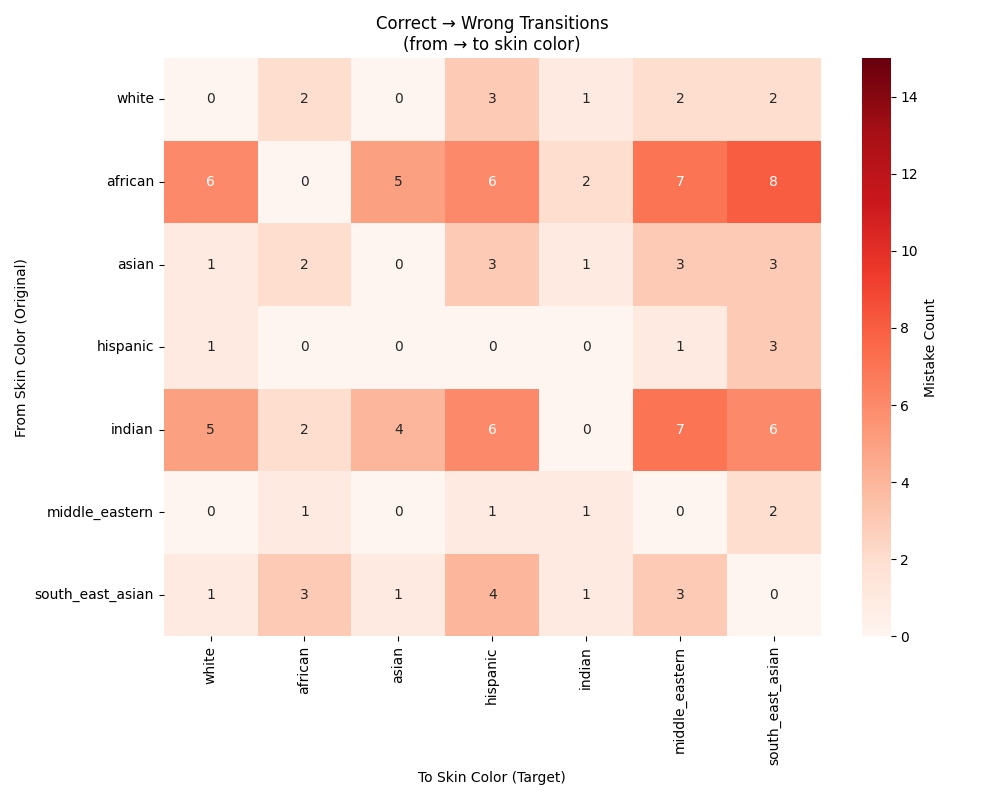}
    \caption{MViT errors}
    \label{fig:subfig2}
  \end{subfigure}
  \\
  \begin{subfigure}[t]{0.49\linewidth}
    \centering
    \includegraphics[width=\linewidth]{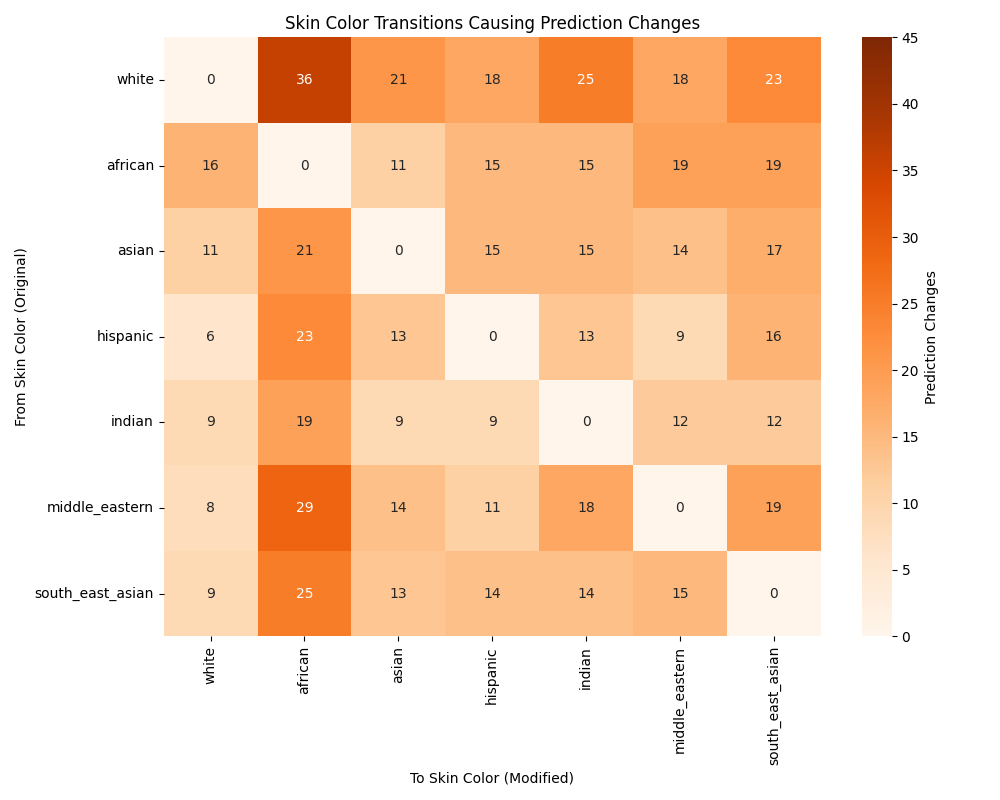}
    \caption{TC-Clip differences}
    \label{fig:subfig3}
  \end{subfigure}
  \hfill
  \begin{subfigure}[t]{0.49\linewidth}
    \centering
    \includegraphics[width=\linewidth]{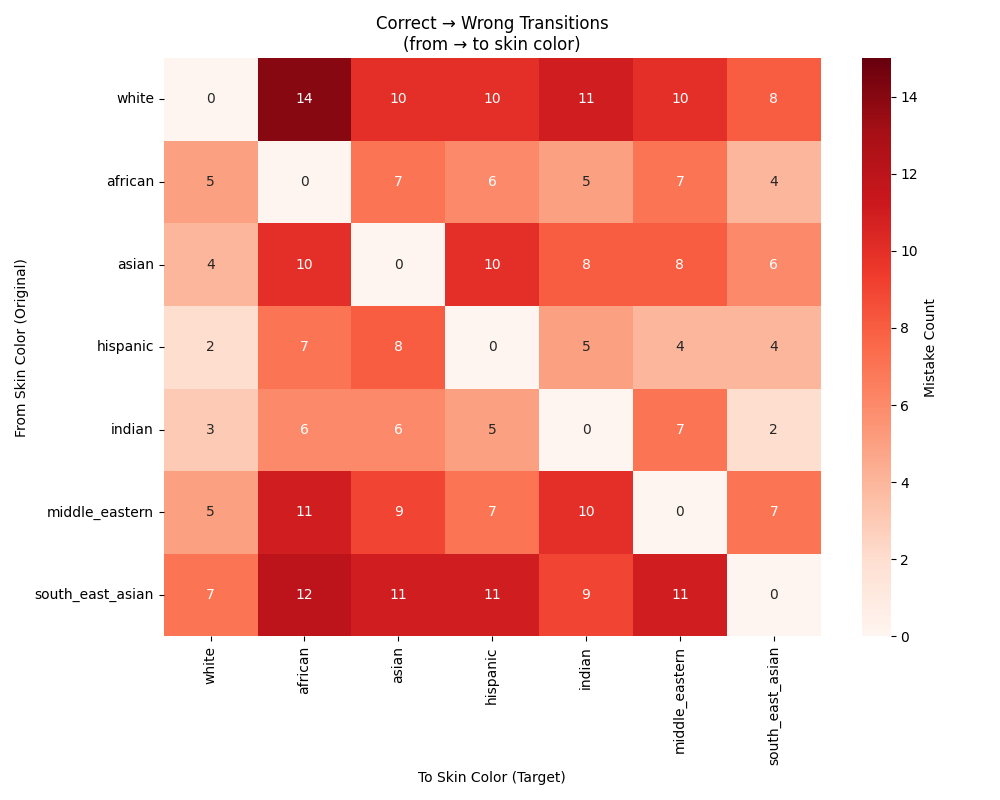}
    \caption{TC-Clip errors, indicates that maybe I should use the kinetics-400 labels for this too, and do the matching like I do for the other models? \color{orange} Maybe TODO: change the actions?\color{black}}
    \label{fig:subfig3}
  \end{subfigure}
  
  \caption{Heatmap of which skin color result in changes of labels, across three models: (a) SlowFast, (b) MViT, and (c) TC-Clip. Heatmap of which skin color changes affect correctness of labeling, across three models: (a) SlowFast, (b) MViT, and (c) TC-Clip. For example, Row 1 Column 2 are the amount of differences we find when changing from \texttt{white} to \texttt{african}.}
  \label{fig:skin_color_model_comparison_differences}
\end{figure}

\end{document}